\documentclass[letterpaper]{article} 
\usepackage{aaai24}  
\usepackage{times}  
\usepackage{helvet}  
\usepackage{courier}  
\usepackage[hyphens]{url}  
\usepackage{graphicx} 
\usepackage{natbib}  
\usepackage{caption} 
\usepackage{algorithm}
\usepackage{algorithmic}

\usepackage{comment}
\usepackage{pifont}
\usepackage{multirow}
\usepackage{amsmath,amssymb,amsfonts,bbding}
\usepackage{color}
\usepackage{colortbl}
\usepackage{xcolor}

\usepackage{newfloat}
\usepackage{listings}
\DeclareCaptionStyle{ruled}{labelfont=normalfont,labelsep=colon,strut=off} 
\floatstyle{ruled}
\newfloat{listing}{tb}{lst}{}
\floatname{listing}{Listing}
\title{Mono3DVG: 3D Visual Grounding in Monocular Images}
\author{
    Yang Zhan\textsuperscript{\rm 1},
    Yuan Yuan\textsuperscript{\rm 1},
    Zhitong Xiong\textsuperscript{\rm 2}
}
\affiliations{
    \textsuperscript{\rm 1}School of Artificial Intelligence, Optics and Electronics (iOPEN), Northwestern Polytechnical University, Xi'an, China\\
    \textsuperscript{\rm 2}Technical University of Munich (TUM), Munich, Germany\\
    \{zhanyangnwpu, y.yuan1.ieee, xiongzhitong\}@gmail.com
    
}

\usepackage{bibentry}

\begin{document}

\maketitle

\begin{abstract}
    We introduce a novel task of 3D visual grounding in monocular RGB images using language descriptions with both appearance and geometry information. Specifically, we build a large-scale dataset, Mono3DRefer, which contains 3D object targets with their corresponding geometric text descriptions, generated by ChatGPT and refined manually. 
    To foster this task, we propose Mono3DVG-TR, an end-to-end transformer-based network, 
    which takes advantage of both the appearance and geometry information in text embeddings for multi-modal learning and 3D object localization.
    Depth predictor is designed to explicitly learn geometry features. 
    The dual text-guided adapter is proposed to refine multiscale visual and geometry features of the referred object.
    Based on depth-text-visual stacking attention, the decoder fuses object-level geometric cues and visual appearance into a learnable query.
    Comprehensive benchmarks and some insightful analyses are provided for Mono3DVG. Extensive comparisons and ablation studies show that our method significantly outperforms all baselines.
    The dataset and code will be released.
\end{abstract}

\section{Introduction}
\label{sec:introduction}

For intelligent systems and robots, understanding objects based on language expressions in real 3D scenes is an important capability for human-machine interaction. Visual grounding \cite{Deng_2021_ICCV, yang2022improving, Zhan10056343} has made significant progress in 2D scenes, but these approaches cannot obtain the true 3D extent of the objects. Therefore, recent researches \cite{chen2020scanrefer,achlioptas2020referit3d} utilize RGB-D sensors for 3D scanning and build indoor point cloud scenes for 3D visual grounding. The latest work \cite{lin2023wildrefer} focuses on outdoor service robots and utilizes LiDAR and an industrial camera to capture point clouds and RGB images as multimodal visual inputs. However, the practical application of these works is limited due to the expensive cost and device limitations of RGB-D scans and LiDAR scans.

Monocular 3D object detection \cite{Huang_2022_CVPR, brazil2023omni3d} can obtain the 3D coordinates of all objects in the scene and only requires RGB images. While this approach has broad applications, it overlooks the semantic understanding of the 3D space and its objects, making it unable to accomplish specific object localization based on human instructions. To carry out more effective human-machine interaction on devices equipped with cameras, such as drones, surveillance systems, intelligent vehicles, and robots, it is necessary to perform visual grounding using natural language in monocular RGB images.

\begin{table*}[h]
\centering
\scalebox{0.8}{
\begin{tabular}{lccccccccc}
\hline
Dataset & Publication &{\begin{tabular}[c]{@{}c@{}}Expression\\ Num.\end{tabular}} & {\begin{tabular}[c]{@{}c@{}}Object\\ Num.\end{tabular}} & {\begin{tabular}[c]{@{}c@{}}Scene\\ Num.\end{tabular}} & Range  &{\begin{tabular}[c]{@{}c@{}}Exp.\\ Length\end{tabular}}  &  Vocab  & Scene   & Target          \\ 
\hline
SUN-Spot    & \textit{ICCVW'2019}    & 7,990                                                          & 3,245                                                              & 1,948                                                               & --        &  14.04  & 2,690   & Indoor           & furni.                    \\
REVERIE     & \textit{CVPR'2020}    & 21,702                                                          & 4,140                                                             & 90                                                                  & --        &  18.00  & 1,600   & Indoor           & furni.                   \\
ScanRefrer  & \textit{ECCV'2020}  & 51,583                                                         & 11,046                                                             & 704                                                                 & 10m       &  20.27 & 4,197  & Indoor           & furni.                    \\
Sr3d  & \textit{ECCV'2020}         &\ \,83,572$^*$                                                          & 8,863                                                             & 1,273                                                               & 10m       & --  &  196 & Indoor           & furni.                    \\
Nr3d    &  \textit{ECCV'2020}       & 41,503                                                          & 5,879                                                             & 642                                                                 & 10m       &  11.40  & 6,951   & Indoor           & furni.                   \\
SUNRefer    &  \textit{CVPR'2021}    & 38,495                                                          & 7,699                                                             & 7,699                                                               & --         &  16.30  & 5,279   & Indoor           & furni.                    \\
STRefer    & \textit{arXiv'2023}     & 5,458                                                          & 3,581                                                              & 662                                                                 & 30m    &   --  &  --   & Outdoor          & human                     \\
LifeRefer    & \textit{arXiv'2023}   & 25,380                                                         & 11,864                                                             & 3,172                                                               & 30m       &  --  & --    & In/Outdoor       & human                    \\
\rowcolor{blue!7} 
\textbf{Mono3DRefer}   & -- & \textbf{41,140}                                                 & \textbf{8,228}                                                    & \textbf{2,025}                                                      & \textbf{102m}   &  \textbf{53.24}  &  \textbf{5,271}  & \textbf{Outdoor} & \textbf{human, vehicle}      \\ 
\hline
\end{tabular}}
\caption{Statistic comparison of visual grounding datasets in the 3D scene, where 'num.' denotes number, 'exp.' indicates expression, and 'furni.' means furniture. '*' represents the unique text data automatically generated and the largest amount.}
\label{datasets}
\end{table*}

\begin{table*}[h]
\centering
\scalebox{0.8}{
\begin{tabular}{lcccccc}
\hline
\multirow{2}{*}{Dataset} & \multicolumn{2}{c}{Language Context} & \multicolumn{2}{c}{Visual Context} & \multirow{2}{*}{Label} & \multirow{2}{*}{Task}  \\ 
\cline{2-5}
                                  & Form           & Cost       & Form        & Cost        &    &                               \\  
\hline
SUN-Spot                        & manual                  & {\color{red}{\FiveStar}{\FiveStar}{\FiveStar}{\FiveStar}{\FiveStar}}                  & RGB-D                & {\color{red}{\FiveStar}{\FiveStar}{\FiveStar}}{\color{red}{\FiveStarOpen}{\FiveStarOpen}}                  & 2D bbox    &  2D Visual Grounding in RGBD                       \\
REVERIE               & manual                  & {\color{red}{\FiveStar}{\FiveStar}{\FiveStar}{\FiveStar}{\FiveStar}}                  & pc                   & {\color{red}{\FiveStar}{\FiveStar}{\FiveStar}{\FiveStar}}{\color{red}{\FiveStarOpen}}                  & 2D bbox       & Localise Remote Object                   \\
ScanRefrer                      & manual                  & {\color{red}{\FiveStar}{\FiveStar}{\FiveStar}{\FiveStar}{\FiveStar}}                  & pc                   & {\color{red}{\FiveStar}{\FiveStar}{\FiveStar}{\FiveStar}}{\color{red}{\FiveStarOpen}}                  & 3D bbox         & 3D Visual Grounding                     \\
Sr3d                 & templated               & {\color{red}{\FiveStar}}{\color{red}{\FiveStarOpen}{\FiveStarOpen}{\FiveStarOpen}{\FiveStarOpen}}                 & pc                   & {\color{red}{\FiveStar}{\FiveStar}{\FiveStar}{\FiveStar}}{\color{red}{\FiveStarOpen}}                  & 3D bbox     &  3D Visual Grounding                       \\
Nr3d         & manual                  & {\color{red}{\FiveStar}{\FiveStar}{\FiveStar}{\FiveStar}{\FiveStar}}                  & pc                   & {\color{red}{\FiveStar}{\FiveStar}{\FiveStar}{\FiveStar}}{\color{red}{\FiveStarOpen}}                  & 3D bbox         &  3D Visual Grounding                   \\
SUNRefer             & manual                  & {\color{red}{\FiveStar}{\FiveStar}{\FiveStar}{\FiveStar}{\FiveStar}}                  & RGB-D                & {\color{red}{\FiveStar}{\FiveStar}{\FiveStar}}{\color{red}{\FiveStarOpen}{\FiveStarOpen}}                  & 3D bbox        &  3D Visual Grounding in RGBD                  \\
STRefer                        & manual                  & {\color{red}{\FiveStar}{\FiveStar}{\FiveStar}{\FiveStar}{\FiveStar}}                  & pc \& RGB            & {\color{red}{\FiveStar}{\FiveStar}{\FiveStar}{\FiveStar}{\FiveStar}}                  & 3D bbox       &  3D Visual Grounding in the Wild                   \\
LifeRefer                     & manual                  & {\color{red}{\FiveStar}{\FiveStar}{\FiveStar}{\FiveStar}{\FiveStar}}                  & pc \& RGB            & {\color{red}{\FiveStar}{\FiveStar}{\FiveStar}{\FiveStar}{\FiveStar}}                & 3D bbox            &  3D Visual Grounding in the Wild                \\
\rowcolor{blue!7} 
\textbf{Mono3DRefer}                & \textbf{ChatGPT+manual}        & {\color{red}{\FiveStar}{\FiveStar}}{\color{red}{\FiveStarOpen}{\FiveStarOpen}{\FiveStarOpen}}         & \textbf{RGB}         & {\color{red}{\FiveStar}}{\color{red}{\FiveStarOpen}{\FiveStarOpen}{\FiveStarOpen}{\FiveStarOpen}}          & \textbf{2D/3D bbox}   &   3D Visual Grounding in RGB               \\     
\hline
\end{tabular}
}
\caption{The form, cost, and label of the datasets collected in Table \ref{datasets} and the corresponding tasks. 'pc' denotes point cloud and 'bbox' means bounding box.}
\label{datasets_taskscost}
\end{table*}

In this work, we introduce a task of 3D object localization through language descriptions with geometry information directly in a single RGB image, termed Mono3DVG (see Fig. \ref{fig:Mono3DVG}). Specifically, we build a large-scale dataset, Mono3DRefer, which provides 41,140 natural language expressions of 8,228 objects. 
Mono3DRefer's descriptions contain both appearance and geometry information, generated by ChatGPT and refined manually.
Geometry information can provide more precise instructions and identify invisible objects. Even if the appearance of an object is the primary visual perception for humans, they tend to use geometry information to distinguish objects.

To perform inference based on the language with appearance and geometry information, we propose a novel end-to-end transformer-based approach, namely Mono3DVG-TR, which consists of a multi-modal feature encoder, a dual text-guided adapter, a grounding decoder, and a grounding head. 
First, we adopt transformer and CNN to extract textual and multi-scale visual features. Depth predictor is designed to explicitly learn geometry features.
Second, to refine multiscale visual and geometry features of the referred object, we propose the dual text-guided adapter to perform text-guided feature learning based on pixel-wise attention.
Finally, a learnable query first aggregates the initial geometric features, then enhances text-related geometric features by text embedding and finally collects appearance features from multiscale visual features. 
The depth-text-visual stacking attention fuses object-level geometric cues and visual appearance into the query, fully realizing text-guided decoding.

Our contributions can be summarized as follows:

\begin{itemize}
\item We introduce a novel task of 3D visual grounding in monocular RGB images using descriptions with appearance and geometry information, termed Mono3DVG.
\item We contribute a large-scale dataset, which contains 41,140 expressions generated by ChatGPT and refined manually based on the KITTI, named Mono3DRefer.
\item We propose an end-to-end transformer-based network, Mono3DVG-TR, which fully aggregates the appearance and geometry features in multi-modal embedding.
\item We provide sufficient benchmarks based on two-stage and one-stage methods. Extensive experiments show that our method significantly outperforms all baselines.
\end{itemize}

\section{Related Work}
\label{relatedwork}

\begin{figure*}[h]
  \centering
  \includegraphics[width=0.95\textwidth]{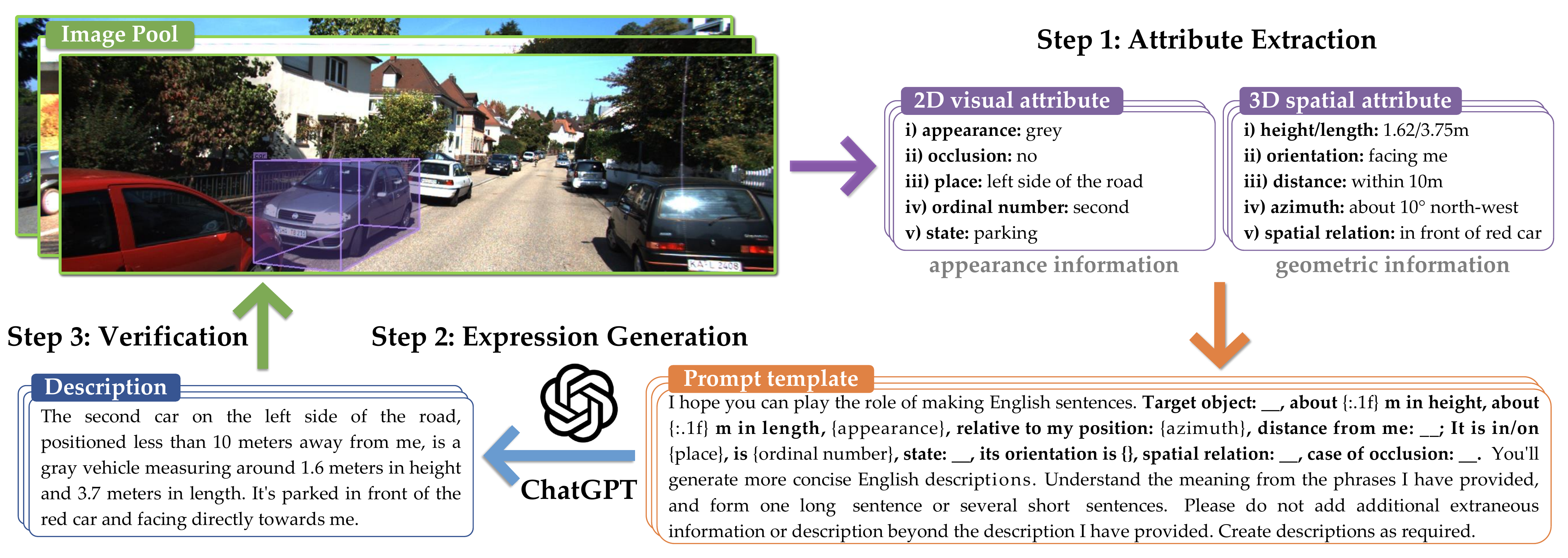}
  \caption{Our data collection pipeline: 
  i) 2D visual attributes that provide appearance information and 3D spatial attributes that provide geometric information of the target are extracted. 
  ii) fill in the prompt template we designed with attributes, and input the complete prompt into ChatGPT to get descriptions. 
  iii) check whether the description can uniquely identify the object.
  }
  \label{fig:dataCollect}
\end{figure*}

\subsection{2D Visual Grounding}
2D visual grounding aims to localize the 2D bounding box of a referred object in an image by a language expression. 
The earlier two-stage approaches \cite{zhang2018grounding, hu2017modeling, yu2018mattnet,liu2019improving,yu2018rethinking,chen2017query} adopt a pre-trained detector to generate region proposals and extract visual features. It obtains the optimal proposal by calculating scores with vision-language features and sorting. 
Additionally, NMTree \cite{liu2019learning} and RvG-Tree \cite{hong2022learning} utilize tree networks by parsing the expression. To capture objects' relation, graph neural network is adopted by \citet{yang2019dynamic,wang2019neighbourhood, yang2020relationship}. 
Recently, the one-stage pipeline has been widely used due to its low computational cost. Many works \cite{chen2018real,sadhu2019zero, yang2019fast, yang2020improving, huang2021look,liao2022progressive} use visual and text encoders to extract visual and textual features, and then fuse the multi-modal features to regress box coordinates. They do not depend on the quality of pre-generated proposals. 
\citet{du2021visual} and \citet{Deng_2021_ICCV} first design the end-to-end transformer-based network, which has achieved superior results in terms of both speed and performance. \cite{li2021referring,sun2022proposal} propose the multi-task framework to further improve the performance. \cite{yang2022improving,ye2022shifting} focus on adjusting visual features by multi-modal features.
\citet{mauceri2019sun} present dataset for 2D visual grounding in RGB-D images. \citet{qi2020reverie} study 2D visual grounding for language-guided navigation in indoor scenes.
However, these works cannot obtain the true 3D coordinates of the object in the real world, which greatly limits the application.

\subsection{Monocular 3D Object Detection}
Monocular 3D object detection aims to predict the 3D bounding boxes of all objects in an image. The methods can be summarized into anchor-based, keypoint-based, and pseudo-depth based methods. 
The anchor-based method requires preset 3D anchors and regresses a relative offset. M3D-RPN \cite{brazil2019m3d} is an end-to-end network that only requires training a 3D region proposal network.
Kinematic3D \cite{brazil2020kinematic} improves M3D-RPN by utilizing 3D kinematics to extract scene dynamics.
Furthermore, some researchers predict key points and then estimate the size and location of 3D bounding boxes, such as SMOKE \cite{liu2020smoke}, FCOS3D \cite{wang2021fcos3d}, MonoGRNet \cite{qin2019monogrnet}, and MonoFlex \cite{zhang2021objects}. However, due to the lack of depth information, pure monocular approaches have difficulty accurately localizing targets.
Other works \cite{8897727, ding2020learning, park2021pseudo, chen2022pseudo} utilize extra depth estimators to supplement depth information.
However, existing models only extract spatial relationships and depth information from visual content. Hence, we propose to explore the impact of language with geometry attributes on 3D object detection.

\subsection{3D Visual Grounding}
3D visual grounding task aims to localize the 3D bounding box of a referred object in a 3D scene by a language expression.
To handle this task, Scanrefer \cite{chen2020scanrefer} and Referit3D \cite{achlioptas2020referit3d} first create datasets.
Similar to the counterpart 2D task, earlier works adopt the two-stage pipeline which uses a pre-trained detector to generate object proposals and extract features, such as PointNet++ \cite{qi2017pointnet++}. 
SAT \cite{yang2021sat} adopts 2D object semantics as extra input to assist training. InstanceRefer \cite{yuan2021instancerefer} converts this task into an instance matching problem.
To understand complex and diverse descriptions in point clouds directly, \citet{feng2021free} construct a language scene graph, a 3D proposal relation graph, and a 3D visual graph.
3DVG-Trans \cite{zhao20213dvg}, TransRefer3D \cite{he2021transrefer3d}, Multi-View Trans \cite{huang2022multi}, and LanguageRefer \cite{roh2022languagerefer} all develop transformer-based  architectures. 
D3Net \cite{chen2022d} and 3DJCG \cite{cai20223djcg} both develop a unified framework for dense captioning and visual grounding.
\citet{liu2021refer} present a novel task for 3D visual grounding in RGB-D images. The previous works are all in indoor environments and target furniture as the object. To promote the application, \citet{lin2023wildrefer} introduce the task in large-scale dynamic outdoor scenes based on online captured 2D images and 3D point clouds.
However, capturing visual data through LiDAR or the industrial camera is expensive and not readily available for a wide range of applications. Our work focuses on the 3D visual grounding in a single image.

\begin{figure*}[h]
  \centering
  \includegraphics[width=0.9\textwidth]{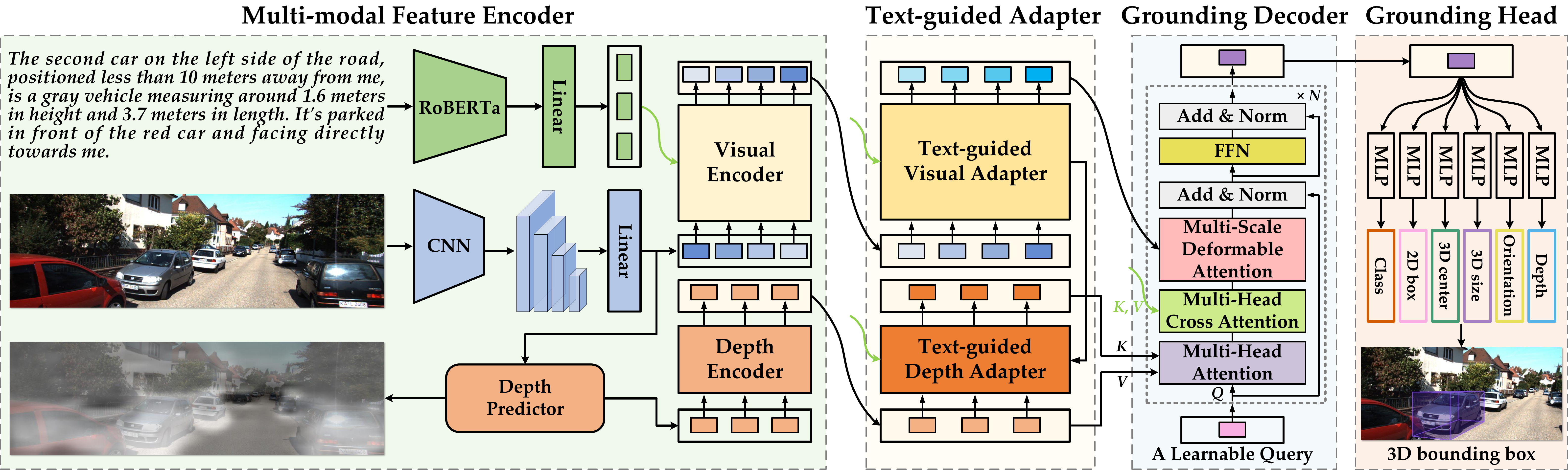}
  \caption{Overview of the proposed framework. 
  The multi-modal feature encoder first extracts textual, multi-scale visual, and geometry features. 
  The dual text-guided adapter refines visual and geometry features of referred objects based on pixel-wise attention.
  A learnable query fuses geometry cues and visual appearance of the object using depth-text-visual stacking attention in the grounding decoder. 
  Finally, the grounding head adopts multiple MLPs to predict the 2D and 3D attributes of the target.
  }
  \label{fig:model}
\end{figure*}

\section{Mono3DRefer Dataset}
\label{sec:dataset}
As shown in Table \ref{datasets} and Table \ref{datasets_taskscost}, previous SUN-Spot \cite{mauceri2019sun} and REVERIE \cite{qi2020reverie} only focus on 2D bounding boxes in the 3D scene. Subsequently, ScanRefer \cite{chen2020scanrefer}, Sr3d, Nr3d \cite{achlioptas2020referit3d}, and SUNRefer \cite{liu2021refer} are built to investigate 3D visual grounding, but they are limited to indoor static scenes. Although STRefer and LifeRefer \cite{lin2023wildrefer} focus on outdoor dynamic scenes, they require LiDAR and industrial cameras. 
To facilitate the broad application of 3D visual grounding, we employ both manually annotated and ChatGPT to annotate a large-scale dataset based on KITTI \cite{geiger2012we} for Mono3DVG.

\subsection{Data Annotation}
\label{sec:dataannotation}
To cover all scenes and reduce inter-frame similarity, we performed scene clustering on the original KITTI dataset and sampled 2025 images from each category. In Fig. \ref{fig:dataCollect}, the annotation pipeline consists of three stages.
\textbf{Step 1: Attribute extraction.}
The attributes of objects are divided into 2D visual attributes (appearance, occlusion, place, ordinal number, state) and 3D spatial attributes (height/length, orientation, distance, azimuth, spatial relationship).
The color of appearance is preliminarily extracted by the HSV color recognition method. Occlusion and height/length are directly obtained from labels of the raw KITTI. Based on the 302 category results of scene clustering, unified rough annotations are performed for the scene place and state of objects in each category. Distance and azimuth are calculated by the coordinates of 3D boxes.
Spatial relations include i) Horizontal Proximity, ii) Between, and iii) Allocentric \cite{achlioptas2020referit3d} such as far from, next to, between A and B, on the left, and in front. The judgment model is established based on 3D boxes and space geometry to preliminarily extract ordinal number, orientation, and spatial relation. 
Finally, to ensure correctness, we organize four people to verify and correct 2D and 3D attributes that provide appearance and geometric information.
\textbf{Step 2: Expression generation.}
We customize the prompt template for generating expressions for ChatGPT. Fill in the template with each attribute of objects and input the complete prompt into ChatGPT to obtain the descriptions.
\textbf{Step 3: Verification.}
To guarantee the correctness of descriptions, four persons from our team jointly verify the dataset.

\subsection{Dataset Statistics}
\label{sec:datastatistic}
Table \ref{datasets} summarizes the statistical information of the dataset. We sample 2025 frames of images from the original KITTI for Mono3DRefer, containing 41,140 expressions in total and a vocabulary of 5,271 words. In addition to the Sr3d generated through templates, Mono3DRefer has a similar number of expressions as the ScanRefer and Nr3d. 
For the range, 10m is the range of the whole scene pre-scanned by RGB-D sensors, 30m is the approximated perception radius with annotations for the LiDAR sensor, and 102m is the distance range of objects with annotations for our dataset.
The average length of expressions generated by ChatGPT is 53.24 words involving visual appearance and geometry information.
Table \ref{datasets_taskscost} shows that the Mono3DVG task has relatively low language data collection costs and the lowest visual data collection costs. 
We provide more detailed statistics and analyses in the supplementary materials.

\section{Methodology}
\label{sec:methods}
As shown in Fig. \ref{fig:model}, we propose an end-to-end transformer-based framework, Mono3DVG-TR, which consists of four main modules: 1) the encoder; 2) the adapter; 3) the decoder; 4) the grounding head.

\subsection{Multi-modal Feature Encoder}
\label{sec:feature}
We leverage pre-trained RoBERTa-base \cite{liu2019roberta} and a linear layer to extract the textual embeddings $\boldsymbol{p}_{t} \in \mathbb{R}^{C \times N_{t}}$, where $N_{t}$ is the length of the input sentence.
For the image $\boldsymbol{I}\in \mathbb{R}^{H \times W\times 3}$, we utilize a CNN backbone (\textit{i.e.}, ResNet-50 \cite{he2016deep} and an additional convolutional layer) and a linear layer to obtain four level multi-scale visual features $\boldsymbol{f}_{v} \in \mathbb{R}^{C \times N_{v}}$, where $C=256$ and $N_{v} = \frac{H}{8} \times \frac{W}{8}+\frac{H}{16} \times \frac{W}{16}+\frac{H}{32} \times \frac{W}{32}+\frac{H}{64} \times \frac{W}{64}$.
Following \citet{zhang2022monodetr}, we use the lightweight depth predictor to get the geometry feature $\boldsymbol{f}_{g} \in \mathbb{R}^{C \times N_{g}}$, where $N_{g}=\frac{H}{16} \times \frac{W}{16}$.
\begin{figure}[t]
  \centering
  \includegraphics[width=0.82\columnwidth]{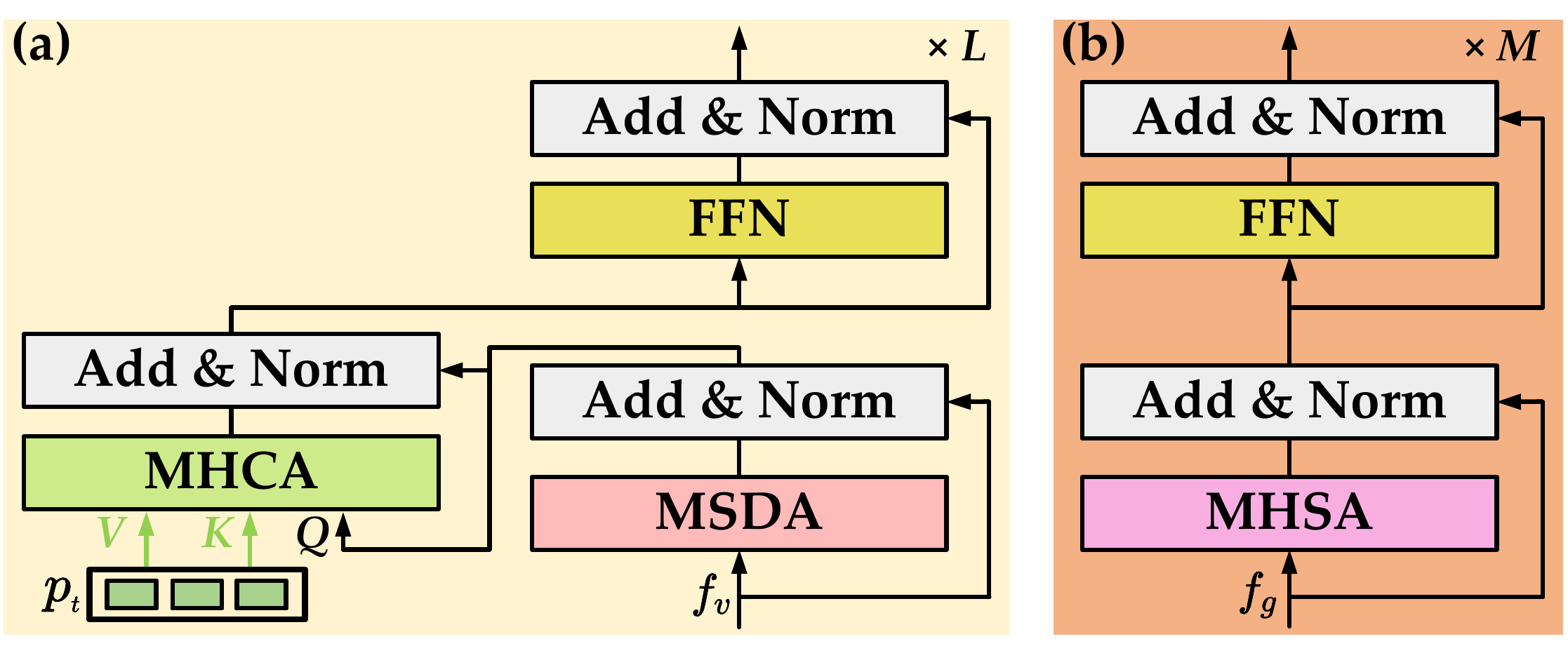}
  \caption{Detail of visual encoder (a) and depth encoder (b).}
  \label{fig:model_encoder}
\end{figure}
Then we design visual encoder and depth encoder to conduct global context inference and generate embeddings with long-term dependencies, denoted as $\boldsymbol{p}_{v} \in \mathbb{R}^{C \times N_{v}}$, $\boldsymbol{p}_{g} \in \mathbb{R}^{C \times N_{g}}$.
The depth encoder is composed of one transformer encoder layer to encode geometry embeddings.
In Fig. \ref{fig:model_encoder}(a), visual encoder replaces multi-head self-attention (MHSA) with multi-scale deformable attention (MSDA) to avoid excessive attention computation on multi-scale visual features. Moreover, we insert an additional multi-head cross-attention (MHCA) layer between MSDA layer and feed-forward network (FFN), providing textual cues for visual embeddings.

\begin{figure}[t]
  \centering
  \includegraphics[width=0.82\columnwidth]{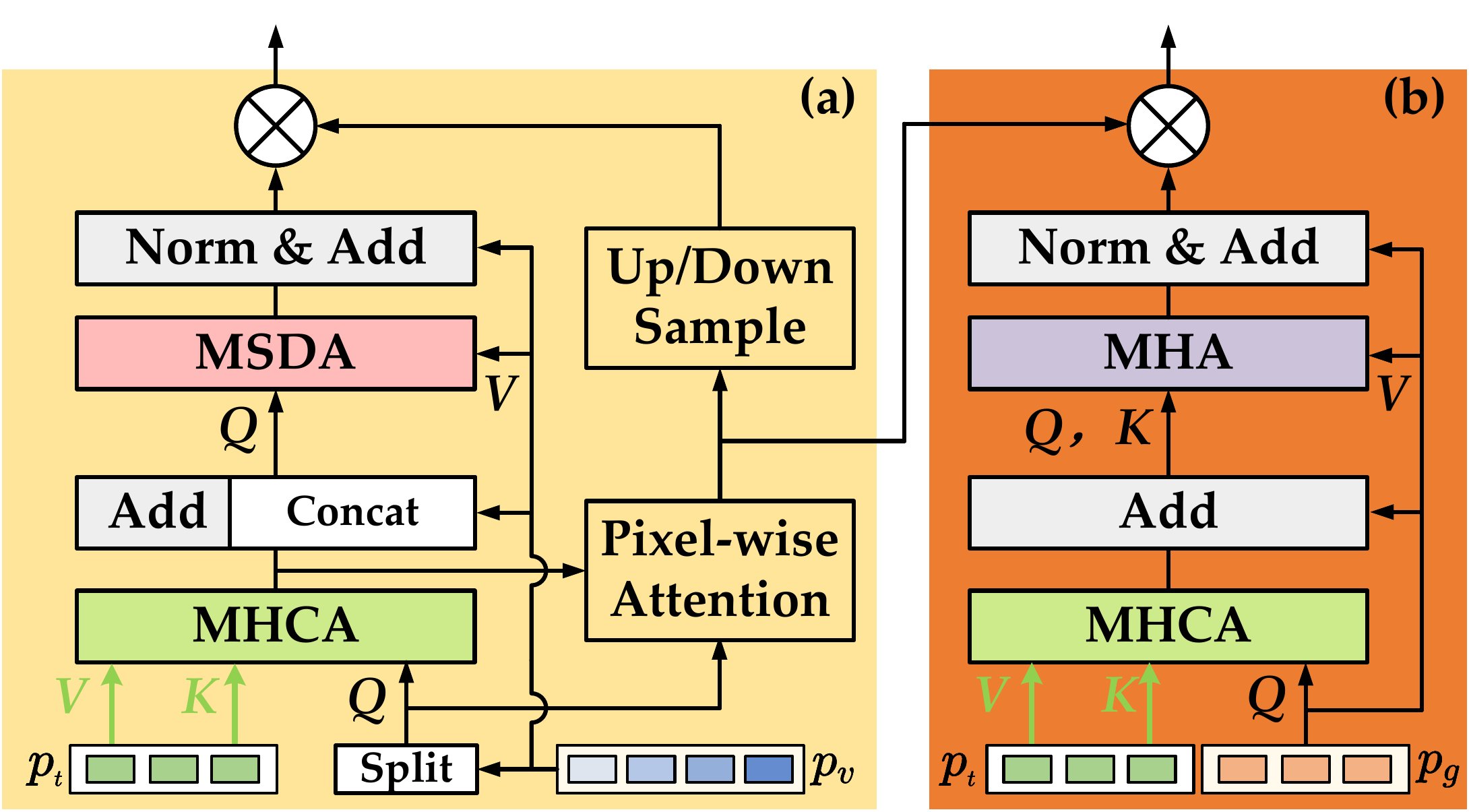}
  \caption{Detail of text-guided visual and depth adapter.}
  \label{fig:model_adapter}
\end{figure}

\subsection{Dual Text-guided Adapter}
\label{sec:Adapter}
To exploit the appearance and geometry information in text, the dual adapter is proposed.
In Fig. \ref{fig:model_adapter}(b), the depth adapter takes the geometry embedding $\boldsymbol{p}_{g}$ as the query for MHCA and takes the text embedding $\boldsymbol{p}_{t}$ as the key and value. 
Then, a multi-head attention (MHA) layer is used to apply implicit text-guided self-attention to the geometry features. 
Original geometry embedding $\boldsymbol{p}_{g}$ as the value. The refined geometry feature is denoted as $\boldsymbol{p}_{g}^{''}$.
Visual adapter requires splitting and concatenating multi-scale visual embeddings $\boldsymbol{p}_{v}$ before and after MHCA which uses $\boldsymbol{p}_{v}^{\frac{1}{16} }$ with the size of $\frac{H}{16} \times \frac{W}{16}$ as the query. Then, MSDA is used instead of MHA, and the refined visual feature is denoted as $\boldsymbol{p}_{v}^{''}$.

Then, we linearly project $\boldsymbol{p}_{v}^{\frac{1}{16} }$ and the output of MHCA in the visual adapter to obtain the original visual feature map $\boldsymbol{F}_{orig} \in \mathbb{R}^{C \times \frac{H}{16} \times \frac{W}{16}}$ and the text-related $\boldsymbol{F}_{text} \in \mathbb{R}^{C \times \frac{H}{16} \times \frac{W}{16}}$, respectively. 
To explore the alignment relationship and fine-grained correlation between vision and language, we compute the attention score $\boldsymbol{s}_{ij} \in \mathbb{R}^{\frac{H}{16} \times \frac{W}{16}}$ for each region ($i, j$) in the feature map as follows:
\begin{equation}
\boldsymbol{F}_{orig}=\left \| \boldsymbol{F}_{orig} \right \| _{2},\text{ } \boldsymbol{F}_{text}=\left \| \boldsymbol{F}_{text} \right \| _{2} ,
\end{equation}
\begin{equation}
\boldsymbol{a} _{ij}^{c}  =\boldsymbol{F}_{orig}^{c}(i,j)\odot \boldsymbol{F}_{text}^{c}(i,j),  c=1,2,\dots ,C  
\end{equation}
\begin{equation}
\boldsymbol{s}_{ij} = \sum_{c=1}^{C}\boldsymbol{a}_{ij}^{c} .
\end{equation}
where, $\left \| \cdot   \right \| _{2} $ and $\odot$ indicate $l_{2}$-norm and element-wise product respectively.
Then, we further model the semantic similarity $\boldsymbol{S}^{\frac{1}{16}}$ with the size of $\frac{H}{16} \times \frac{W}{16}$ between each pixel feature and the text feature using the Gaussian function:
\begin{equation}
\boldsymbol{S}^{\frac{1}{16}}  =\alpha \cdot \text{exp} ( -\frac{ ( 1- \boldsymbol{s}_{ij} )^{2}  }{2\sigma^{2} }  ) ,
\end{equation}
where, $\alpha$ and $\sigma $ are a scaling factor and standard deviation, respectively, and both are learnable parameters. 
We upsample $\boldsymbol{S}^{\frac{1}{16}}$ using bilinear interpolation and downsample $\boldsymbol{S}^{\frac{1}{16}}$ using max pooling. Then we concatenate the flattened score maps to obtain the multi-scale attention score ${S} \in \mathbb{R}^{N_{v}}$:

\begin{equation}
S = \text{Concat}[\text{Up}(\boldsymbol{S}^{\frac{1}{16}}),\boldsymbol{S}^{\frac{1}{16}} ,\text{Down}(\boldsymbol{S}^{\frac{1}{16}}) ,\text{Down}(\boldsymbol{S}^{\frac{1}{16}})].
\end{equation}

\begin{table*}[h]
\centering
\scalebox{0.9}{
\begin{tabular}{lcccccccc}
\hline
\multirow{2}{*}{{Method}} &\multirow{2}{*}{{Type}} & \multicolumn{2}{c}{{Unique}}  & \multicolumn{2}{c}{{Multiple}} & \multicolumn{2}{c}{{Overall}} & Time cost\\ 
 &   & {Acc@0.25} & {Acc@0.5} & {Acc@0.25}  & {Acc@0.5} & {Acc@0.25} & {Acc@0.5} & (ms)\\ 
 \hline
CatRand   & Two-Stage  & \underline{\textcolor{gray}{100}}  & \underline{\textcolor{gray}{100}}    & 24.47    & 24.43    & 38.69   & 38.67  &0  \\
Cube R-CNN + Rand  & Two-Stage  & 32.76 &14.61 & 13.36 & 7.21 & 17.02  & 8.60 &153\\
Cube R-CNN + Best  & Two-Stage  & 35.29 &16.67 & 60.52 & 32.99 & 55.77 & 29.92&153 \\ \hline
ZSGNet + backproj  & One-Stage & 9.02 & 0.29 & 16.56  & 2.23  & 15.14 & 1.87&31\\
FAOA + backproj   & One-Stage  & 11.96  & 2.06 & 13.79 & 2.12 & 13.44 & 2.11&144 \\
ReSC + backproj   & One-Stage & 11.96   & 0.49 & 23.69 & 3.94 & 21.48 & 3.29&97\\
TransVG + backproj & Tran.-based & 15.78  & 4.02 & 21.84& 4.16  & 20.70 & 4.14 &80\\
\rowcolor{blue!7} 
Mono3DVG-TR (Ours) & Tran.-based  & \textbf{57.65} & \textbf{33.04} & \textbf{65.92}  & \textbf{46.85}  & \textbf{64.36}& \textbf{44.25} &110\\ 
\hline
\end{tabular}
}
\caption{
Comparison with baselines. The underline means performance exceeding our bolded results.
}
\label{Mono3DVG}
\end{table*}

Based on pixel-wise attention scores, the visual and geometry features are focused on the regions relevant to the textual description.
We use the features $ p_{v}^{''}$ and $ p_{g}^{''}$ and scores ($S^{\frac{1}{16}} \in \mathbb{R}^{N_{d}}$ is flattened) to perform element-wise multiplication, resulting in adapted features of the referred object:
\begin{equation}
    \tilde{\boldsymbol{p}} _{v} =\boldsymbol{p}_{v}^{''} \cdot S  ,\text{ } \tilde{\boldsymbol{p}} _{g} =\boldsymbol{p}_{g}^{''}  \cdot S^{\frac{1}{16}} .
\end{equation}

\subsection{Grounding Decoder}
\label{sec:Decoder}
As shown in Fig. \ref{fig:model}, the $n$-th decoder layer consists of a block composed of MHA, MHCA, and MSDA, and an FFN. 
The learnable query $\boldsymbol{p}_{q} \in \mathbb{R}^{C \times 1}$ first aggregates the initial geometric information, then enhances text-related geometric features by text embedding, and finally collects appearance features from multi-scale visual features. 
This depth-text-visual stacking attention adaptively fuses object-level geometric cues and visual appearance into the query.

\subsection{Grounding Head}
\label{sec:Head}
Our grounding head employs multiple MLPs for 2D and 3D attribute prediction. 
The output of the decoder, \textit{i.e.}, the learnable query, is denoted by $\tilde{\boldsymbol{p}}_{q} \in \mathbb{R}^{C \times 1}$. 
Then, $\tilde{\boldsymbol{p}}_{q}$ is separately fed into a linear layer for predicting the object category, a 3-layer MLP for the 2D box size ($l,r,t,b$) and projected 3D box center ($x_{3D},y_{3D}$), a 2-layer MLP for the 3D box size ($h_{3D},w_{3D},l_{3D}$), a 2-layer MLP for the 3D box orientation $\theta $, and a 2-layer MLP for the depth $d_{reg}$. 
($l,r,t,b$) represents the distances between the four sides of the 2D box and the projected 3D center point ($x_{3D},y_{3D}$). 
Similar to \cite{zhang2022monodetr}, the final predicted depth $d_{pred}$ is computed.

\subsection{Loss Function}
\label{sec:loss}
We group the category, 2D box size, and projected 3D center as 2D attributes, and the 3D box size, orientation, and depth as 3D attributes. The loss for 2D is formulated as: 
\begin{equation}\label{eq:loss2d}
    \mathcal{L}_{2D} = \lambda_{1} \mathcal{L}_{class} + \lambda_{2} \mathcal{L}_{lrtb} + \lambda_{3} \mathcal{L}_{GIoU} + \lambda_{4} \mathcal{L}_{xy3D} ,
\end{equation}
where, $\lambda_{1\sim4}$ is set to $(2,5,2,10)$ following \cite{zhang2022monodetr}.
$\mathcal{L}_{class}$ is Focal loss \cite{lin2017focal} for predicting nine categories. $\mathcal{L}_{lrtb}$ and $\mathcal{L}_{xy3D}$ adopt the L1 loss. $\mathcal{L}_{GIoU}$ is the GIoU loss \cite{rezatofighi2019generalized} that constrains the 2D bounding boxes. The loss for 3D is defined as: 
\begin{equation}\label{eq:loss3d}
    \mathcal{L}_{3D} = \mathcal{L}_{size3D} + \mathcal{L}_{orien}  + \mathcal{L}_{depth}. 
\end{equation}
We use the 3D IoU oriented loss \cite{Ma_2021_CVPR}, MultiBin loss \cite{Chen_2020_CVPR}, and Laplacian aleatoric uncertainty loss \cite{Chen_2020_CVPR} as $\mathcal{L}_{size3D}$, $\mathcal{L}_{orien}$, and $\mathcal{L}_{depth}$ to optimize the predicted 3D size, orientation, and depth. Following \cite{zhang2022monodetr}, we use Focal loss to supervise the prediction of the depth map, denoted as $\mathcal{L}_{dmap}$. Finally, our overall loss is formulated as: 
\begin{equation}\label{eq:loss}
    \mathcal{L}_{overall} =  \mathcal{L}_{2D}+ \mathcal{L}_{3D} + \mathcal{L}_{dmap}.
\end{equation}

\section{Experiments}
\label{sec:experiments}

\textbf{Implementation Details}. 
We split our dataset into 29,990, 5,735, and 5,415 expressions for train/val/test sets respectively.
We train 60 epochs with a batch size of 10 by AdamW with $10^{-4}$ learning rate and $10^{-4}$ weight decay on one GTX 3090 24-GiB GPU. The learning rate decays by a factor of 10 after 40 epochs.
The dropout ratio is set to 0.1.

\textbf{Evaluation metric}. 
Similar to \cite{chen2020scanrefer, liu2021refer, lin2023wildrefer}, we use the accuracy with 3D IoU threshold (Acc@0.25 and Acc@0.5) as our metrics, where the threshold includes 0.25 and 0.5.

\textbf{Baselines}. 
To explore the difficulty and enable fair comparisons, we design several baselines and validate these methods using a unified standard. 
\textbf{Two-stage:} 
1) CatRand randomly selects a ground truth box that matches the object category as the prediction result. This baseline measures the difficulty of our task and dataset. 2) (Cube R-CNN \cite{brazil2023omni3d} + Rand) randomly selects a bounding box that matches the object category as the prediction result from predicted object proposals of Cube R-CNN, the best monocular 3D object detector. 
3) (Cube R-CNN \cite{brazil2023omni3d} + Best) selects a bounding box that best matches the ground truth box from predicted object proposals. This baseline provides the upper bound on how well the two-stage approaches work for our task.
\textbf{One-stage:}
2DVG backproj baselines adapt the results of 2D visual grounding to 3D by using back-projection. We select three SOTA one-stage methods, \textit{i.e.}, ZSGNet \cite{sadhu2019zero}, FAOA \cite{yang2019fast} , ReSC \cite{yang2020improving}, and the transformer-based TransVG \cite{Deng_2021_ICCV}.

\begin{table*}[h]
\centering
\scalebox{0.9}{
\begin{tabular}{lccccccc}
\hline
\multirow{2}{*}{{Method}} & \multirow{2}{*}{{Type}} & \multicolumn{2}{c}{{Near / Easy}}  & \multicolumn{2}{c}{{Medium / Moderate}} & \multicolumn{2}{c}{{Far / Hard}} \\
                     &             & {Acc@0.25} & {Acc@0.5} & {Acc@0.25}  & {Acc@0.5} & {Acc@0.25} & {Acc@0.5} \\ 
                                 \hline
CatRand            & Two-Stage         & 31.16/47.29                  & 31.05/47.26                 & 35.49/33.92                   & 35.49/33.92                 & \underline{52.11}/30.83                  & \underline{52.11}/\underline{30.74}                 \\
Cube R-CNN + Rand    & Two-Stage       & 17.40/21.12                  & 11.45/11.41                 & 18.01/17.85                   & 8.15/8.01                 & 14.91/10.56                  & 6.38/5.18                 \\
Cube R-CNN + Best   & Two-Stage         & \underline{67.76}/59.66                  & 41.45/33.05                 & 60.69/60.56                   & 30.35/33.45                 & 34.72/46.25                  & \underline{17.01}/22.52                 \\ \hline
ZSGNet + backproj    & One-Stage       & 24.87/21.33                  & 0.59/3.35                 & 16.74/13.87                   & 3.71/0.63                 & 2.15/7.57                  & 0.07/0.84                 \\
FAOA + backproj      & One-Stage    & 18.03/17.51                  & 0.53/3.43                 & 15.64/12.18                   & 3.95/1.34                 & 4.86/8.83                  & 0.62/0.90                 \\
ReSC + backproj  & One-Stage       & 33.68/27.90                  & 0.59/5.71                 & 24.03/19.23                   & 6.15/1.97                 & 4.24/14.41                  & 1.25/1.02                 \\
TransVG + backproj  & Tran.-based        & 29.34/28.88                  & 0.86/6.95                 & 25.05/16.41                   & 8.02/2.75                 & 4.17/12.91                  & 0.97/1.38                 \\
\rowcolor{blue!7} 
Mono3DVG-TR (Ours)               & Tran.-based     & \textbf{64.74}/\textbf{72.36}                 & \textbf{53.49}/\textbf{51.80}                &  \textbf{75.44}/\textbf{69.23}                 & \textbf{55.48}/\textbf{48.66}                 & \textbf{45.07}/\textbf{49.01}                  & \textbf{15.35}/\textbf{29.91}                \\  
\hline
\end{tabular}
}
\caption{
Results for 'near'-'medium'-'far' subsets and 'easy'-'moderate'-'hard' subsets. The underline means performance exceeding our bolded results.
}
\label{Mono3DVG_depth}
\end{table*}

To analyze the importance of other information besides the category, we report metrics of these baselines on 'unique' and 'multiple' subsets in Table \ref{Mono3DVG}.
The 'unique' subset means cases where there is one object that matches the category, while the 'multiple' subset contains multiple confused objects with the same category.
To analyze the task difficulty, we report metrics at varying levels of depth $d$ as near: 0 $< d  \le$ 15m, medium: 15m $< d \le$ 35m, far: 35m $< d \le \infty $ in Table~\ref{Mono3DVG_depth}.
Considering that occlusion or truncation of the objects adds challenge to the task, we also show metrics at varying levels of difficulty as easy: no occlusion and truncation $<$ 0.15, moderate: no/partial occlusion and truncation $<$ 0.3, hard: others.
For more convincing results, we show the average of 5 evaluations with different random seeds for CatRand and Cube R-CNN Rand.

\begin{figure*}
  \centering
  \includegraphics[width=0.985\textwidth]{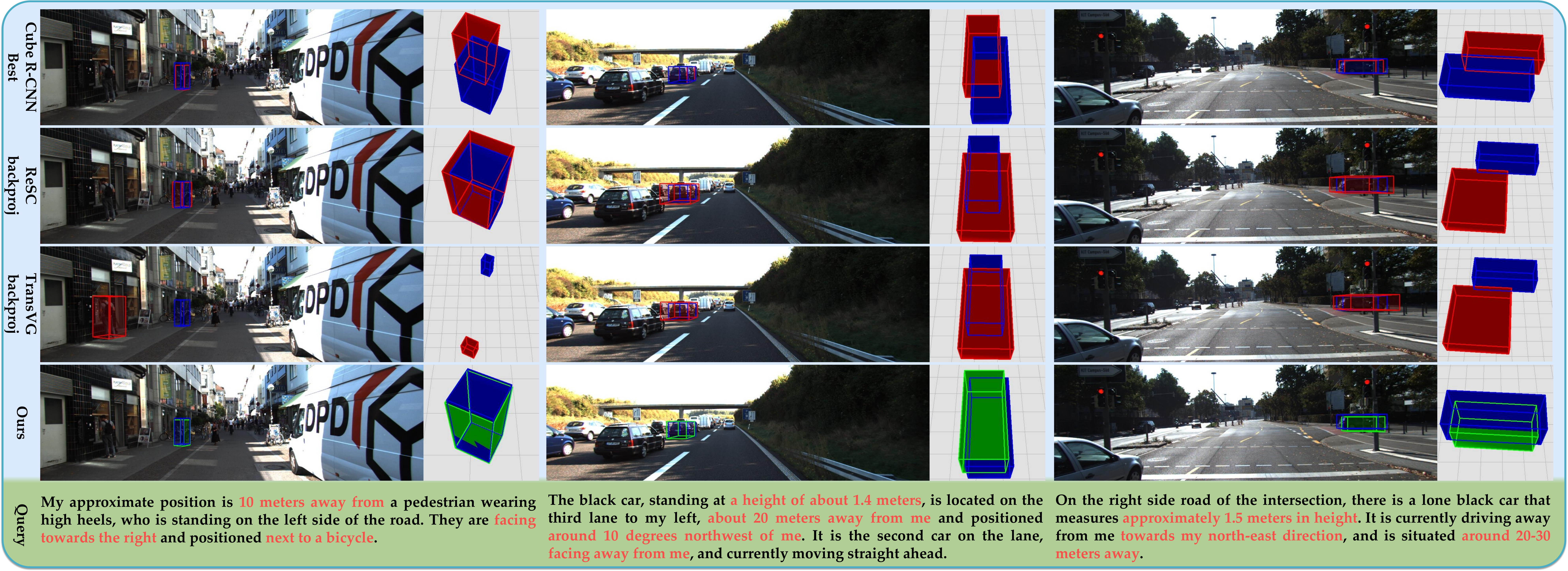}
  \caption{Qualitative results from baseline methods and our Mono3DVG-TR. \textcolor{blue}{Blue}, \textcolor{green}{green}, and \textcolor{red}{red} boxes denote the ground truth, prediction with IoU higher than 0.5, and prediction with IoU lower than 0.5, respectively.}
  \label{fig:result_show}
\end{figure*}

\subsection{Quantitative Analysis and Task Difficulty}
In Table \ref{Mono3DVG}, CatRand achieves 100\% accuracy on the 'unique' subset but only 24\% on the 'multiple'. Cube R-CNN Rand also performs better on the 'unique' subset compared to the 'multiple'. If there is only one car in an image, inputting the "car" is sufficient. However, if there are multiple cars, additional information beyond the category is necessary. 
The significant gap between Cube R-CNN Best and CatRand on the 'unique' subset indicates tremendous research potential in monocular 3D object detection. Overall, while our result is close to the CatRand, there is still room for improvement.

In Table \ref{Mono3DVG_depth}, CatRand performs much better on the 'far' subset compared to 'near' and 'medium'. Our method and other baselines show a decreasing performance as the depth increases. The 'far' subset contains fewer ambiguous objects, so CatRand's random selection of ground truth can achieve better results. Other methods rely on predicted bounding boxes. Generally, objects that are farther away from the camera are more challenging to accurately predict their depth and 3D extent. 
Cube R-CNN Best exhibits excellent results on Acc@0.25.
The accuracy gap between CatRand and our method on the 'far' subset indicates that accurately predicting the depth of target objects based on a single image and natural language is a challenge in our task.

For 'easy-moderate-hard' subsets, Cube R-CNN Best has suboptimal results on Acc@0.25, but a lower Acc@0.5, indicating that the best object detector has the ability to detect occluded or truncated objects, but the accuracy needs to be improved. Our method fully fuses visual and textual features to accurately detect occluded and truncated objects, achieving better results than CatRand. 

Our method outperforms all 2DVG backproj baselines by a significant margin in Tables \ref{Mono3DVG}-\ref{Mono3DVG_depth}. 
It is inefficient to obtain accurate 3D bounding boxes from 2D localization results by back projection. The methods of 2DVG can only predict the extent of the object in the 2D plane and lack the ability to estimate depth, resulting in inaccurate 3D localization.

\subsection{Qualitative Analysis}
\label{sec:qualitative}
Fig. \ref{fig:result_show} displays the 3D localization results of Cube R-CNN Best, ReSC backproj, TransVG backproj, and our proposed method. 
Although the approximate range of objects can be obtained, Cube R-CNN Best fails to provide precise bounding boxes. ReSC backproj and TransVG backproj depend on the accuracy of 2D boxes and are unable to estimate depth, thus unable to provide accurate 3D bounding boxes.
Our method includes text-RGB and text-depth two branches to make full use of the appearance and geometry information for multi-modal fusion, but there are also some failures.
We provide more detailed analyses in the supplementary.

\subsection{Ablation Studies}
\label{sec:ablation}
We conduct detailed ablation studies to validate the effectiveness of our proposed network and report the Acc@0.25 and Acc@0.5 overall on the Mono3DRefer test set.
In Table \ref{VGTR_ablation}, we report results of a comprehensive ablation experiment on the main components.
The first row shows the results by directly using visual and geometry features of the CNN backbone and depth predictor to decode. 
The second row shows a significant improvement with the addition of the encoder. 
In the third row, we only utilize the text-guided visual adapter. 
After adding the complete adapter, the results can be improved by approximately 4\%-5\%.
We provide more detailed analyses of ablation studies in the supplementary.

\begin{table}[t]
\centering
\scalebox{0.9}{
\begin{tabular}{c|cc|cc|cc}
\hline
\multirow{2}{*}{\begin{tabular}[c]{@{}c@{}}Grouning\\ Decoder\end{tabular}} &  \multicolumn{2}{c|}{Encoder} & \multicolumn{2}{c|}{Adapter} & \multirow{2}{*}{Acc0.25} & \multirow{2}{*}{Acc@0.5} \\
&       V.    & D.   & V.    & D.   &         &          \\ \hline
\checkmark    &   &   &    &    &  47.31   &  24.38   \\
\checkmark    & \checkmark& \checkmark  &   &   & 60.21  & 38.52 \\
\checkmark   &  \checkmark& \checkmark  &  \checkmark &   &61.98  &40.12  \\
\checkmark   & \checkmark &  \checkmark & \checkmark  & \checkmark  & \textbf{64.36}     & \textbf{44.25}   \\ \hline
\end{tabular}
}
\caption{The ablation studies of the proposed components of our approach. 'V.' and 'D.' denote visual and depth.}
\label{VGTR_ablation}
\end{table}

\section{Conclusion}
\label{sec:conclusion}
We introduce the novel task of Mono3DVG, which localizes 3D objects in RGB images by descriptions.
Notably, we contribute a large-scale dataset, Mono3DRefer, which is the first dataset that leverages the ChatGPT to generate descriptions. We also provide a series of benchmarks to facilitate future research. 
Finally, we hope that Mono3DVG can be widely applied since it does not require strict conditions such as RGB-D sensors, LiDARs, or industrial cameras. 

\appendix
\section{Appendix}
In this supplementary material, we provide additional details on the data statistic and analysis of the Mono3DRefer dataset (Section A); we also provide implementation details of our Mono3DVG-TR network (Section B), as well as additional ablation studies (Section C) and qualitative comparisons (Section D).

\begin{table}[h]
\centering
\begin{tabular}{crrr}
\hline
      & \multicolumn{1}{c}{Unique} & \multicolumn{1}{c}{Multiple} & \multicolumn{1}{c}{Overall} \\ \hline
Train & 5,910 (19.71\%)              & 24,080 (80.29\%)            & 29,990                       \\
Val   & 1,210 (21.10\%)              & 4,525 (78.90\%)             & 5,735                        \\
Test  & 1,020 (18.84\%)              & 4,395 (81.16\%)             & 5,415                        \\
Total & 8,140 (19.79\%)              & 33,000 (80.21\%)            & 41,140                       \\ \hline
\end{tabular}
\caption{Mono3DRefer dataset statistics on the 'unique' and  'multiple' subsets.}
\label{unqi_m_over}
\end{table}

\begin{figure}[h]
  \centering
  \includegraphics[width=0.95\columnwidth]{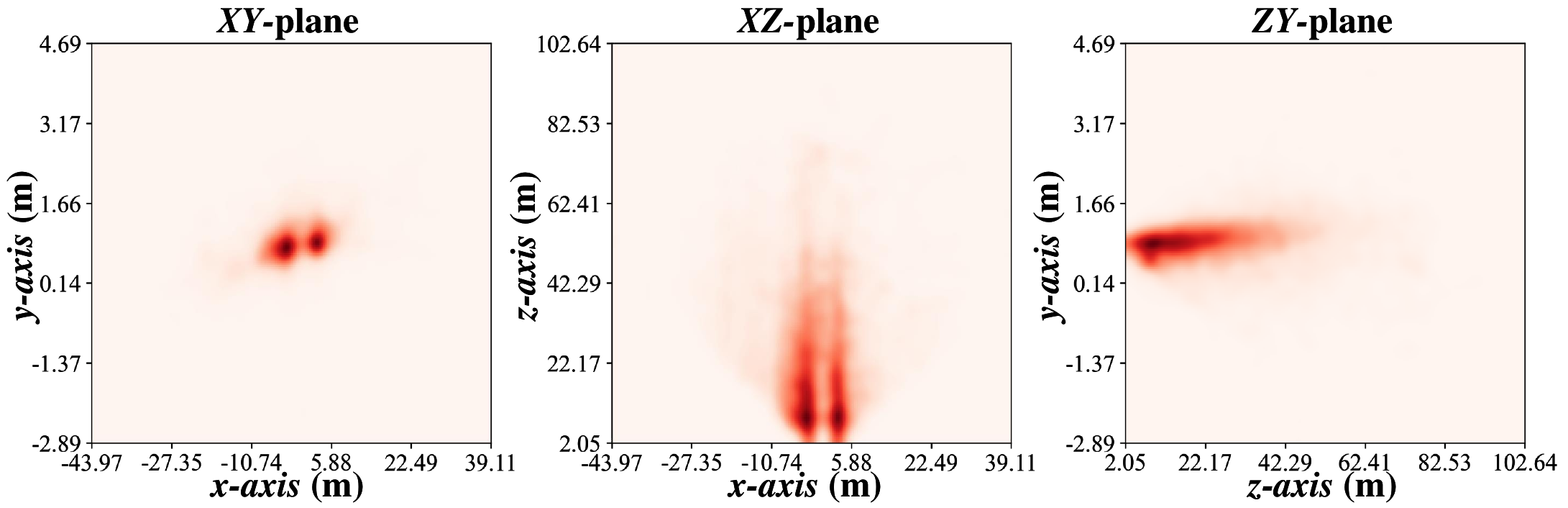}
  \caption{Heatmaps showing the distribution of the 3D object centers on the XY-plane (Front view), XZ-plane (Top view), and ZY-plane (Right view).}
  \label{fig:3DXYZ}
\end{figure}

\subsection{A Dataset}
\subsection{A.1 Dataset Statistics}
We provide Mono3DRefer dataset statistics on the training, validation, and testing in Table \ref{unqi_m_over}. We also provide detailed statistics on the 'unique' and  'multiple' subsets. 
Moreover, the detailed statistics of 'near'-'medium'-'far' subsets and 'easy'-'moderate'-'hard' subsets are shown in Table \ref{near_m_far}.
It can be found that the proportions of different subsets in the training, validation, and testing set are all similar.
As shown in Table \ref{unqi_m_over}, there are more instances in the 'multiple' subset than in the 'unique' subset. In the remaining subsets, only the 'medium' and 'easy' subsets account for more than 40\%, and the proportions of the others are similar.

\begin{table*}[h]
\centering
\begin{tabular}{crrrr}
\hline
      & \multicolumn{1}{c}{Near} & \multicolumn{1}{c}{Medium} & \multicolumn{1}{c}{Far} & \multicolumn{1}{c}{Overall} \\ \hline
Train & 7,805 (26.03\%)        & 12,815 (42.73\%)         & 9,370 (31.24\%)   & 29990 \\   
Val   & 1,575 (27.46\%)        & 2,525 (44.03\%)          & 1,635 (28.51\%)   & 5735  \\  
Test  & 1,520 (28.07\%)        & 2,455 (45.34\%)          & 1,440 (26.59\%)  & 5415  \\  
Total & 10,900 (26.49\%)       & 17,795 (43.25\%)        & 12,445 (30.25\%)  & 41,140        \\ \hline
 & \multicolumn{1}{c}{Easy} & \multicolumn{1}{c}{Moderate} & \multicolumn{1}{c}{Hard} & \multicolumn{1}{c}{Overall} \\ \hline
Train & 13,855 (46.20\%)        & 7,425 (24.76\%)         & 8,710 (29.04\%)   & 29990 \\   
Val   & 2,705 (47.17\%)        & 1,390 (24.24\%)          & 1,640 (28.60\%)   & 5735  \\  
Test  & 2,330 (43.03\%)        & 1,420 (26.22\%)          & 1,665 (30.75\%)  & 5415  \\  
Total & 18,890 (45.92\%)       & 10,235 (24.88\%)        & 12,015 (29.21\%)  & 41,140        \\ \hline
\end{tabular}
\caption{Mono3DRefer dataset statistics on the 'near'-'medium'-'far' subsets and 'easy'-'moderate'-'hard' subsets.}
\label{near_m_far}
\end{table*}

\begin{figure*}[h]
  \centering
  \includegraphics[width=0.85\textwidth]{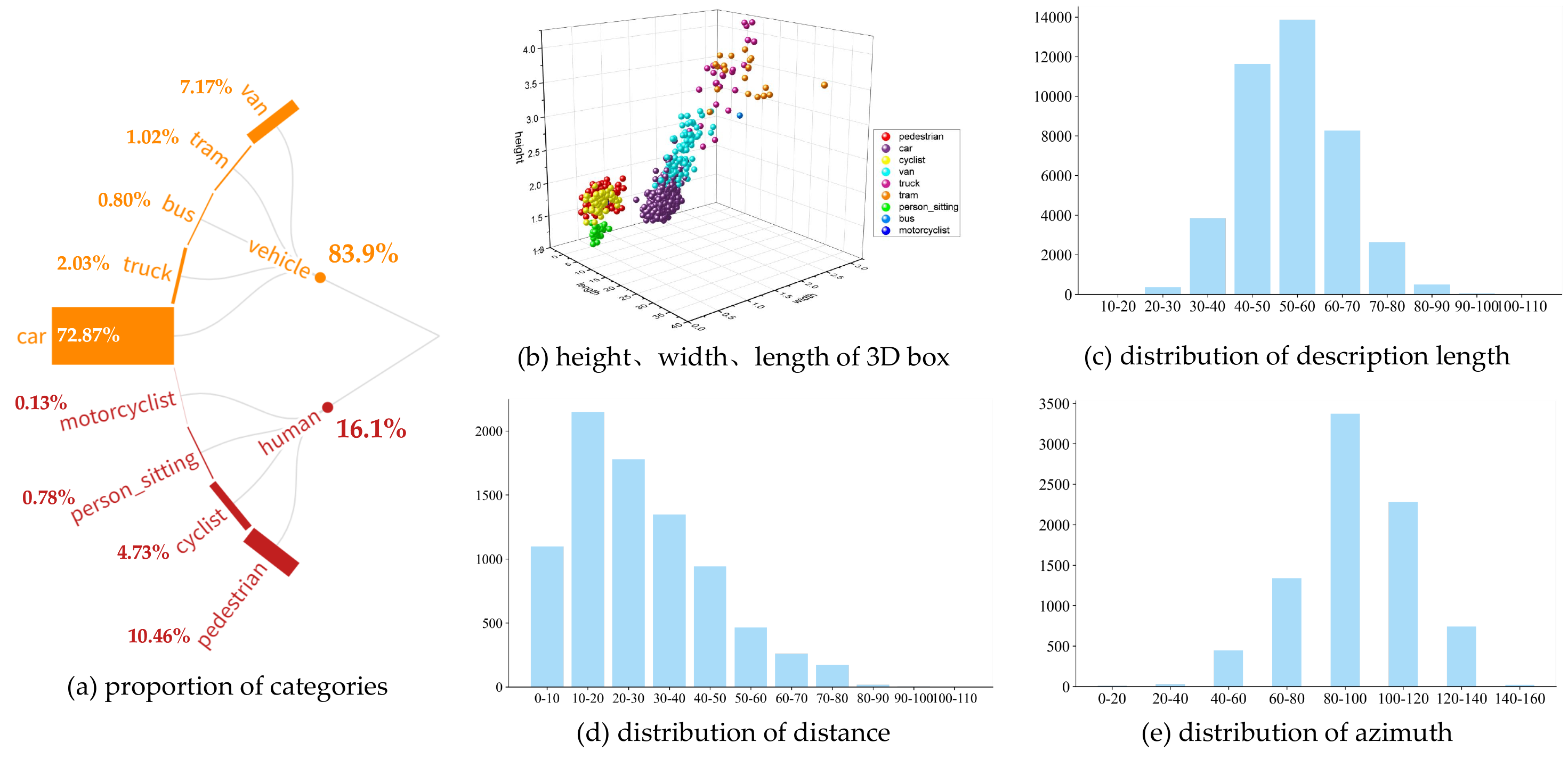}
  \caption{Statistical analysis of the constructed Mono3DRefer.}
  \label{fig:category_word_3DLHW}
\end{figure*}

\subsection{A.2 Dataset Analysis}
\label{sec:dataanalysis}
We present the proportion of each category of the Mono3DRefer dataset in Fig. \ref{fig:category_word_3DLHW}(a). 
Mono3DRefer mainly focuses on the most common traffic elements: vehicle and human. To be more fine-grained, vehicle can be further divided into car, truck, bus, tram and van. Meanwhile, human can be divided into pedestrian, cyclist, motorcyclist, and person sitting. Mono3DRefer provides a large coverage of car.
Fig. \ref{fig:category_word_3DLHW}(b) shows the scatter plot of the length, width, and height of the 3D box for each category. It can be seen that the 3D box sizes of different categories of objects are quite different, and the objects of the same category are similar. 
Fig. \ref{fig:category_word_3DLHW}(c) shows the distribution of the description length. Description length refers to the number of words, the shortest query is 14, the longest is 107, and the average length is 53.24.
Fig. \ref{fig:category_word_3DLHW}(d) shows the distribution of the distance. Distance represents the depth of the object. As the distance increases, the number of objects first increases and then decreases.
Fig. \ref{fig:category_word_3DLHW}(e) shows the distribution of the azimuth. Since the target is mainly in front of the camera, the target is mainly concentrated on both sides of the front, that is, within the range of 90°±30°.

\textbf{Layout statistics.} 
Fig. \ref{fig:3DXYZ} illustrates the distribution heatmaps of the projections of 3D object centers on the XY plane, XZ plane, and ZY plane. We found that the objects are distributed approximately 40m on both sides of the camera, with the maximum distance from the camera reaching 102m, indicating a wide area. However, the distribution is more concentrated within a small range.

\begin{figure*}[]
  \centering
  \includegraphics[width=0.95\textwidth]{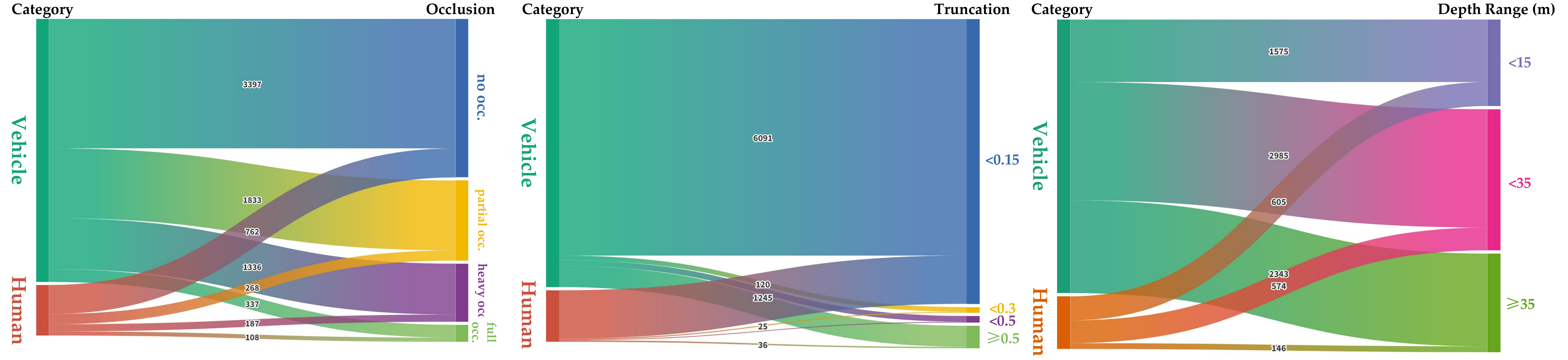}
  \caption{The Sankey diagram of category-occlusion, category-truncation, and category-depth, where “occ.” denotes occlusion.}
  \label{fig:data_trun_occlu_distance}
\end{figure*}

\textbf{Difficulty statistics.}
Fig. \ref{fig:data_trun_occlu_distance} shows the proportion of occlusion, truncation, and depth range of the target in the categories of vehicle and human, respectively. 
In our dataset, the proportion of objects with occlusion is 46.24\%, among which the proportion of partial occlusion and heavy occlusion is similar, and very few objects are fully occluded. The proportion of objects with truncation is 13.16\%. We take 15m and 35m as boundaries to perform interval statistics on the depth range of the object.
Most of the objects are in the medium distance range. The proportion of objects in the near and far distance ranges is similar.

\subsection{B Additional Implementation Details}

\subsection{B.1 Multi-modal Feature Encoder}
The process for depth encoder is:
\begin{equation}
    \begin{aligned}
        f_{d}^{'}& = \text{LN}\left ( f_{d}+  \text{MHSA}( f_{d}  )  \right ), \\
        p_{d} &= \text{LN}\left ( f_{d}^{'}+ \text{FFN}( f_{d}^{'} )  \right ).
    \end{aligned}
\end{equation}
where LN(·) indicates layer normalization.

The process for visual encoder is:
\begin{equation}
    \begin{aligned}
        f_{v}^{'}&= \text{LN}\left ( f_{v}+ \text{MSDA} ( f_{v} ) \right) , \\
        f_{v}^{''}&= \text{LN}\left ( f_{v}^{'}+ \text{MHCA} ( f_{v}^{'}, p_{t}, p_{t}) \right),\\
        p_{v}&= \text{LN}\left ( f_{v}^{''}+ \text{FFN} ( f_{v}^{''} )  \right ).
    \end{aligned}
\end{equation}

\subsection{B.2 Dual Text-guided Adapter}
\label{sec:Adapter}
It is challenging for the decoder to accurately capture target features directly from the original visual and depth embeddings, which often leads to ambiguity during inference. 
To make full use of the appearance and geometry information in the text, we design a dual text-guided adapter with text-RGB and text-depth.
The text-guided visual adapter and depth adapter can aggregate multi-scale visual features and geometry features of the referred objects. The adapted features provide more discriminative text-aware representations for the decoder.

For the depth adapter, the refined geometry features of the referred object are calculated as follows:

\begin{equation}
    \begin{aligned}
         \boldsymbol{p}_{d}^{'} &=  \boldsymbol{p}_{d}+ \text{MHCA} ( \boldsymbol{p}_{d},\boldsymbol{p}_{t}, \boldsymbol{p}_{t} ) ,\\
         \boldsymbol{p}_{d}^{''}&=  \text{LN} (\boldsymbol{p}_{d})  + \text{LN}(\text{MHA} ( \boldsymbol{p}_{d}^{'}, \boldsymbol{p}_{d}^{'},\boldsymbol{p}_{d} )) .
    \end{aligned}
\end{equation}

For the visual adapter, MSDA is used instead of MHA, and the calculation is as follows:
\begin{equation}
    \begin{aligned}
        \boldsymbol{p}_{v}^{\frac{1}{16}'}&= \boldsymbol{p}_{v}^{\frac{1}{16} }+ \text{MHCA} ( \boldsymbol{p}_{v}^{\frac{1}{16} } ,\boldsymbol{p}_{t}, \boldsymbol{p}_{t})  ,\\
        \boldsymbol{p}_{v}^{'}&= \text{Concat}[\boldsymbol{p}_{v}^{\frac{1}{8}}, \boldsymbol{p}_{v}^{\frac{1}{16}'},\boldsymbol{p}_{v}^{\frac{1}{32}},\boldsymbol{p}_{v}^{\frac{1}{64}} ] ,\\
         \boldsymbol{p}_{v}^{''}&=  \text{LN} ( \boldsymbol{p}_{v})  + \text{LN}(\text{MSDA}  ( \boldsymbol{p}_{v}^{'}, \boldsymbol{p}_{v} ))  .
    \end{aligned}
\end{equation}

\subsection{B.3 Grounding Decoder}
\label{sec:Decoder}
For the grounding encoder, the computation is formulated as:
\begin{equation}
\begin{aligned}
    p_{q}^{n'} &= \text{LN}\left ( p_{q}^{n} + \text{MSDA}\left ( \text{MHCA}\left ( \text{MHA}\left ( p_{q}^{n} \right )  \right )  \right )   \right )  ,\\
    p_{q}^{n+1}&= \text{LN}\left ( p_{q}^{n'} + \text{FFN}\left ( p_{q}^{n'} \right )  \right )  .
\end{aligned}
\end{equation}

\subsection{C Additional Runtime Study}
\label{sec:runtime}
We compared the inference time cost (ms) of different baselines in Table \ref{Mono3DVG}. The time cost of our method is not minimal, because our method not only needs to deal with visual language multi-modality but also needs to make depth prediction.
Refer to WildRefer \cite{lin2023wildrefer}, our approach is still efficient and has the potential for real-time applications.

\subsection{D Additional Ablation Studies}
\label{sec:ablation}
We conducted detailed ablation experiments for the visual encoder, depth encoder, and decoder and report the Acc@0.25 and Acc@0.5 of overall on the Mono3DRefer test set.

\begin{table}[h]
\centering
\begin{tabular}{ccc}
\hline
 Architecture & Acc@0.25 & Acc@0.5   \\ \hline
  T $\rightarrow$ D $\rightarrow$ V   &  47.81     &  25.85   \\
  D $\rightarrow$ T $\rightarrow$ V   & \textbf{64.36}     & \textbf{44.25} \\
  D $\rightarrow$ V $\rightarrow$ T   & 53.20        &  25.98  \\
  Full Add\&Norm    &  61.08       & 39.12      \\ \hline
\end{tabular}
\caption{The design of decoder. 'D', 'T', and 'V' represent depth multi-head attention, text multi-head cross attention, and visual multi-scale deformable attention. 
'Full Add\&Norm' means that 'D', 'T', and 'V' are each followed by one Add\&Norm layer in our 'DTV'.}
\label{decoder_design}
\end{table}

\textbf{Decoder Architecture}. 
We experiment with different grounding decoder architectures to make the learnable query better for the fusion of visual appearance and geometry features. 
Based on aggregating geometric features first and then collecting visual features, we stack three attention modules in different orders. 
In Table \ref{decoder_design}, 'D', 'T', and 'V' represent the three attention modules respectively, and the best performance is achieved by the 'D$\rightarrow$T$\rightarrow$V' order. By placing 'D' first, the query can obtain the initial geometric features. Then, the geometric features of the referred objects are enhanced based on the text embeddings while aggregating appearance information. 
Finally, the target appearance features are collected from multi-scale visual features. We believe that stacking three attentions without residual connections and finally adding an FFN can better aggregate the features of referred objects. The fourth row of results represents the stacking attention with the full add\&norm.

\textbf{Visual and Depth Encoder}. 
In Table \ref{encoder_num}, we experiment different layer numbers of visual encoder and depth encoder.
According to the results, we achieved the best performance for the vision encoder L=3 and the depth encoder M=1.

\begin{table}[t]
\centering
\scalebox{0.85}{
\begin{tabular}{c|ccccc}
\hline
    & Num. & Acc0.25 & Acc@0.5 & Params & GFLOPS \\ \hline
\multirow{3}{*}{\begin{tabular}[c]{@{}c@{}}Visual\\ Encoder\end{tabular}} & L = 2             &        62.16 &  43.05     &  118.99M       & 66.39 \\
 & L = 3             &  \textbf{64.36}     & \textbf{44.25} & 119.35M        &  70.09  \\
   & L = 4             &    63.74     & 42.50      &  119.71M       & 73.80 \\ \hline
\multirow{3}{*}{\begin{tabular}[c]{@{}c@{}}Depth\\ Encoder\end{tabular}}  & M = 1             &  \textbf{64.36}     & \textbf{44.25}   & 119.35M        &  70.09   \\
  & M = 2             &  61.26       & 41.75     & 119.48M        & 70.35  \\
 & M = 3             &  60.18       &  38.80     &  119.61M       & 70.60 \\ \hline
\end{tabular}
}
\caption{The different number of encoder layers in visual and depth encoder.}
\label{encoder_num}
\end{table}

\textbf{Decoder Layers}. 
Table \ref{decoder_num} shows the results for different numbers of decoder layers.
Performance is highest when N=1. As the number of decoder layer N increases, the accuracy begins to decline.

\begin{table}[t]
\centering
\begin{tabular}{ccccc}
\hline
 Num. & Acc@0.25 & Acc@0.5 & Params & GFLOPS  \\ \hline
  N = 1   & \textbf{64.36}     & \textbf{44.25}   &  119.35M        &  70.09 \\
  N = 2   & 62.84    &  43.52   & 120.25M         & 70.76\\
  N = 3   & 62.34    & 40.80   &  121.16M        & 71.43 \\
  N = 4   & 60.38    & 38.56    &  122.06M   & 72.10 \\
  N = 5   & 56.14    & 34.28 &  122.97M      & 72.77  \\ \hline
\end{tabular}
\caption{The different numbers of decoder layers used to perform reasoning for Mono3DVG.}
\label{decoder_num}
\end{table}

\subsection{E Additional Qualitative Analysis}
\label{sec:qualitative}
We provide visualization analyses of our method and additional examples of localization results.

\begin{figure*}[h]
  \centering
  \includegraphics[width=0.99\textwidth]{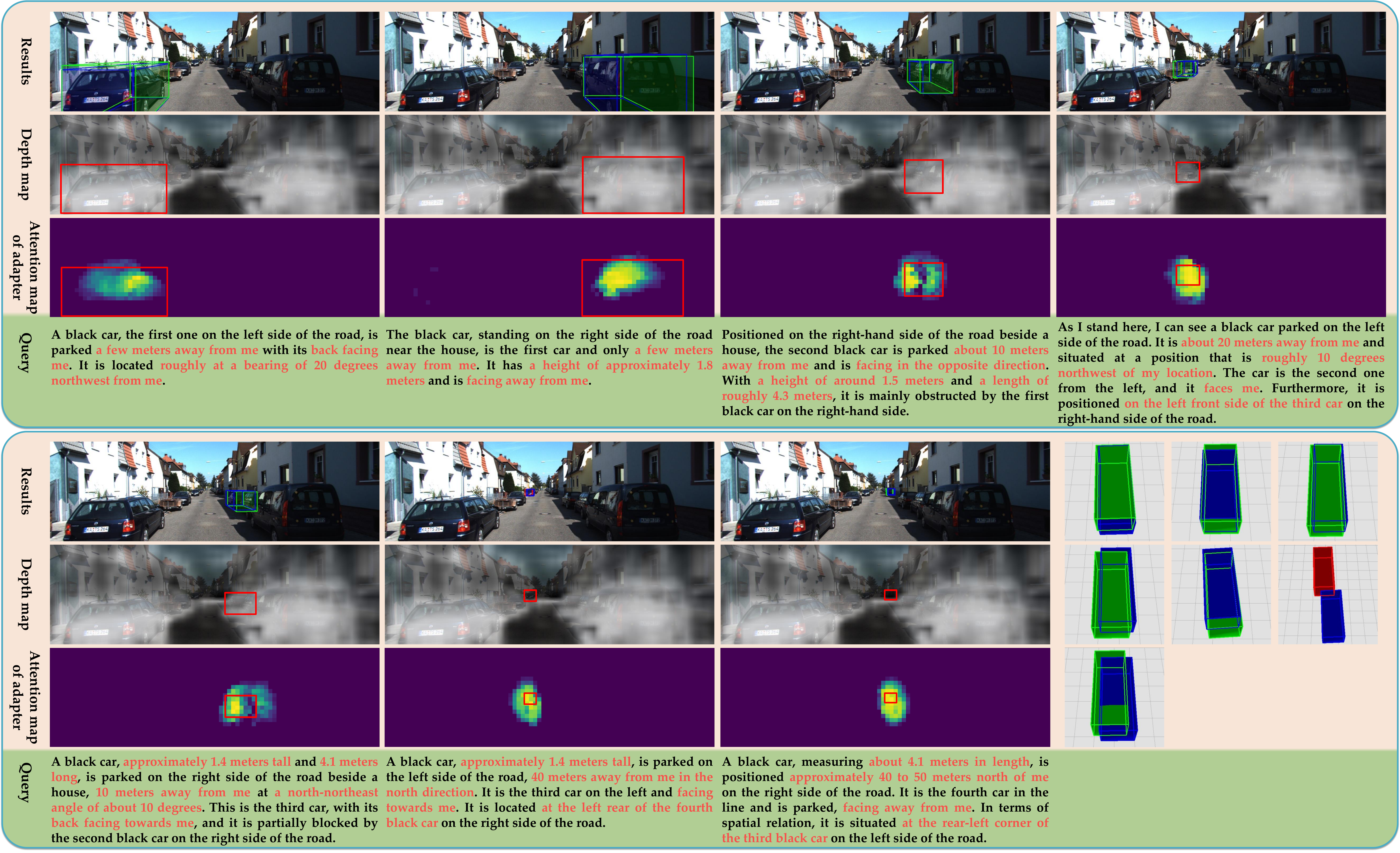}
  \caption{Visualization of '000152.png' image's localization results, the depth predictor's depth maps, and the text-guided adapter's attention score maps for our Mono3DVG-TR.}
  \label{fig:result_maps}
\end{figure*}

\begin{figure*}[h]
    \centering
    \begin{minipage}[t]{0.99\linewidth}
        \centering
        \includegraphics[width=0.99\textwidth]{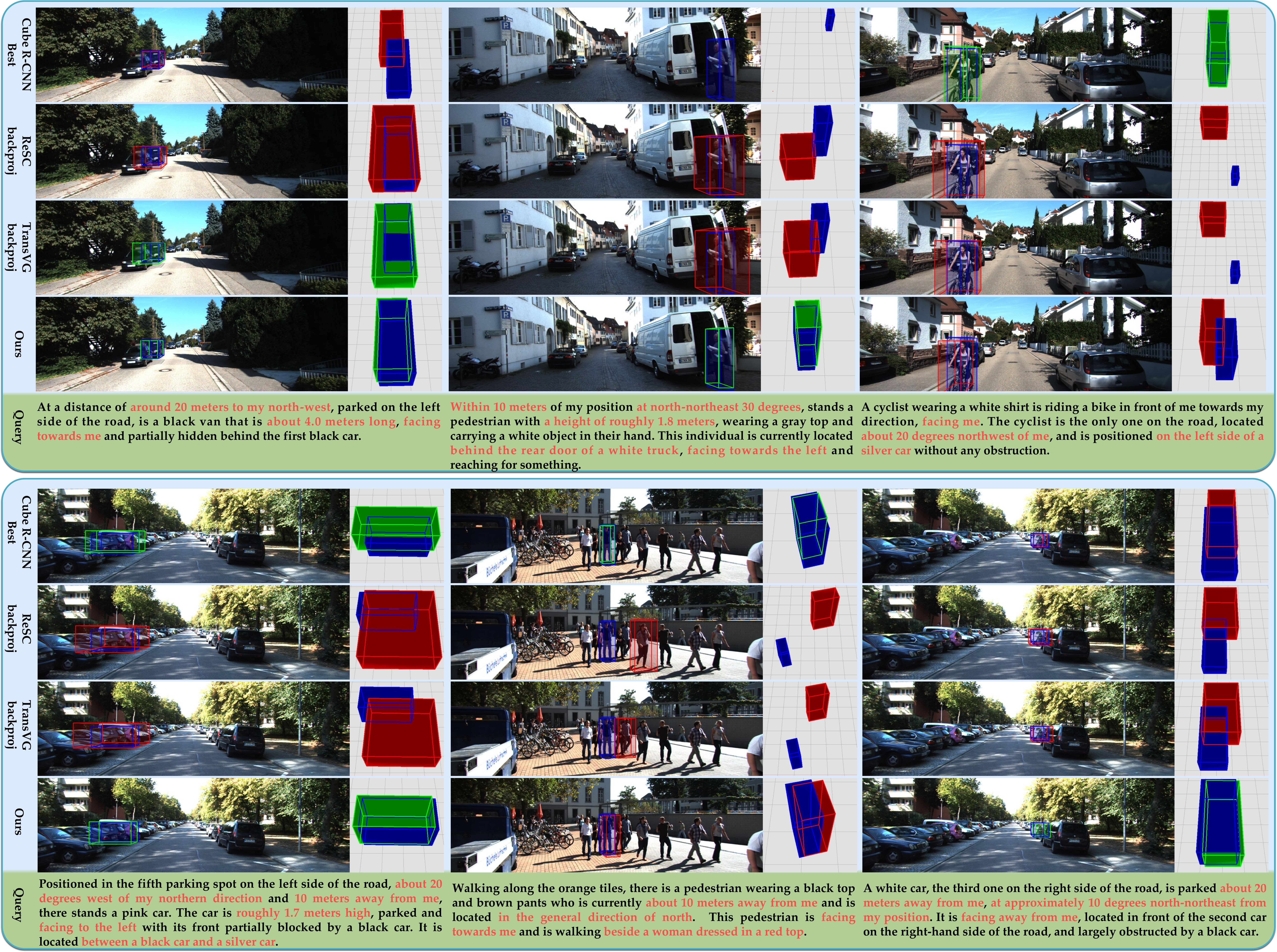}
        \caption{Additional qualitative results in the 'unique'(top) and  'multiple'(bottom) subsets.}
        \label{fig:result_unique}
    \end{minipage}
    \\
    \begin{minipage}[t]{0.99\linewidth}
        \centering
        \includegraphics[width=0.99\textwidth]{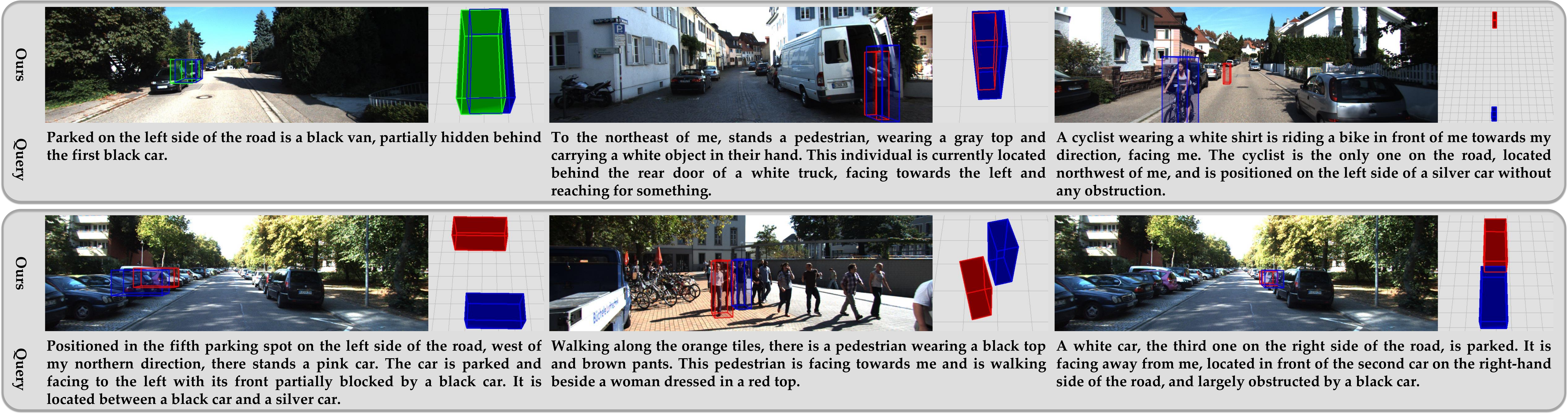}
        \caption{ Qualitative results for the traditional query without specific geometry information corresponding to Fig. \ref{fig:result_unique}.
          }
        \label{fig:result_unique2}
    \end{minipage}
\end{figure*}

\begin{figure*}[h]
  \centering
  \includegraphics[width=0.99\textwidth]{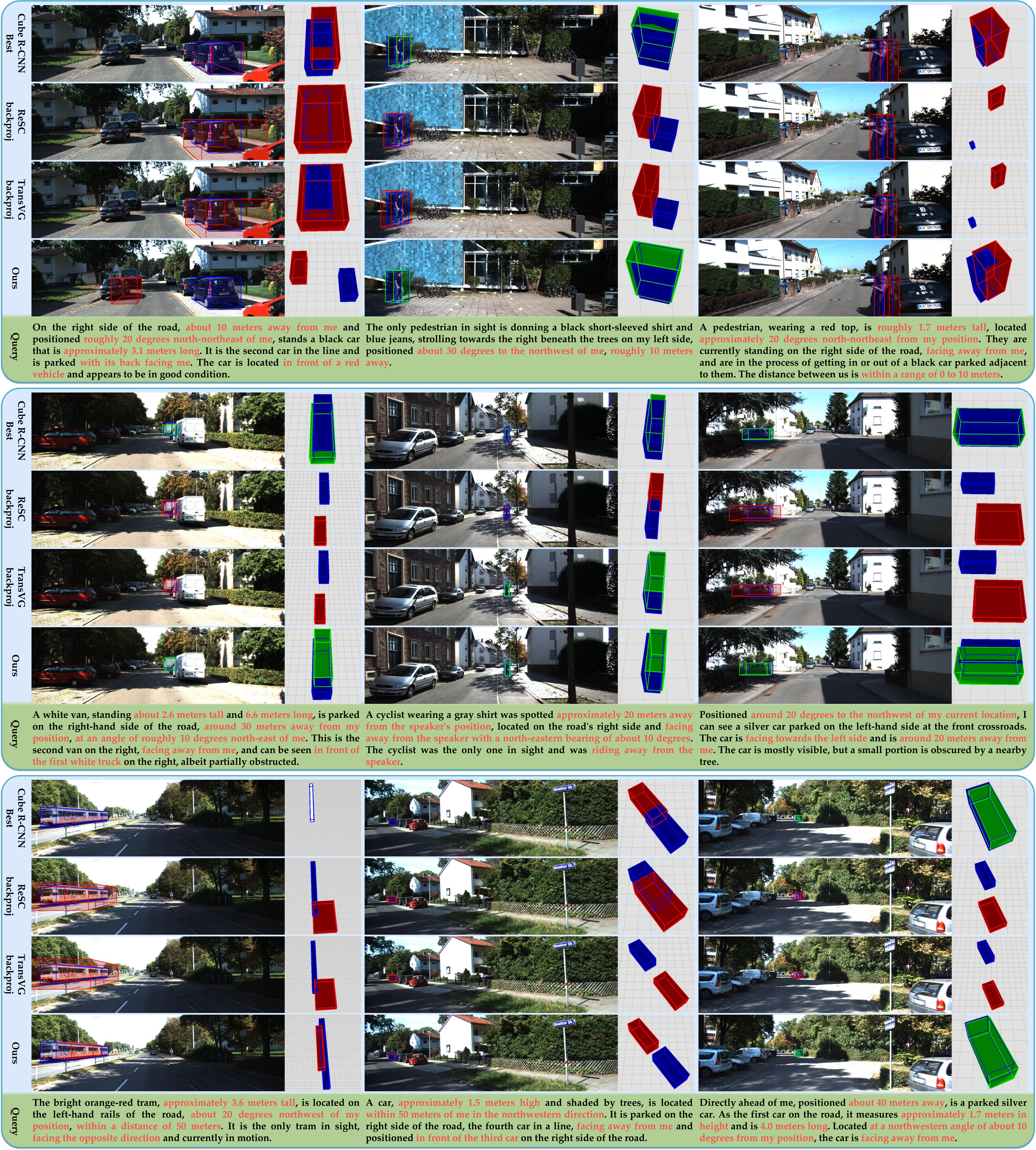}
  \caption{Additional qualitative results in the 'near', 'medium', and 'far' subsets.}
  \label{fig:result_nearfar}
\end{figure*}
\textbf{Visualization analysis of our method.}
In Fig. \ref{fig:result_maps}, we visualize the '000152.png' image's localization results of our Mono3DVG-TR, the depth predictor's depth maps, the text-guided adapter's attention maps, and the grounding decoder's attention maps. The '000152.png' image includes 7 objects.
Observing the foreground depth map in the second row, objects of smaller depths appear white, while larger object depths appear black. The background area mainly appears black. The depth predictor can accurately provide object-wise depths. 
From the attention maps of the adapter, it can be observed that the described target objects generally receive high attention scores in the corresponding regions. The adapted visual and depth features can aggregate in the relevant regions of the target object, filtering out the influence of redundant features. The decoder successfully attends to the target object, achieving accurate 3D localization.

\textbf{More qualitative examples in the 'unique' and 'multiple' subsets.}
To illustrate the difference in performance between the baselines, we provide more qualitative results. We split the results into 'unique' and 'multiple' subsets in Fig. \ref{fig:result_unique}. 
The top and bottom blocks in Fig. \ref{fig:result_unique} represent the results of 'unique' and 'multiple' subsets, respectively.
For the 'unique' subset, the ReSC backproj and TransVG backproj can predict accurate 2D boxes and is able to obtain the rough 3D extent of the object.
In contrast, for the 'multiple' subset, ReSC backproj and TransVG backproj depend on the accuracy of 2D boxes and are unable to estimate object depth, thus unable to localize the object. 
Although ground truth can be obtained to provide accurate target boxes, Cube R-CNN Best sometimes fails to provide targets, such as the second pedestrian in the 'unique' subset.
Our method combines natural language to identify and localize the object in the 'unique' and 'multiple' subsets. Despite some failure cases, it can provide a relatively general 3D range.

\textbf{More qualitative examples in the 'near', 'medium', and 'far' subsets.}
The three blocks in Fig. \ref{fig:result_nearfar} from top to bottom represent the results of 'near', 'medium', and 'far' subsets.
For the 'near' subset, it is not always possible to produce precise localization, but for the majority of cases, a rough 3D box can be obtained. 
In the first case of Fig. \ref{fig:result_nearfar}, our method may fail to handle all spatial relationships and result in ambiguity when facing 'near' targets.
ReSC backproj and TransVG backproj is more likely to predict more accurate 2D boxes in the 'near' subset, thus providing an approximate 3D range. For the 'middle' and 'far' subsets, the ReSC backproj and TransVG backproj may fail due to the lack of depth estimation. 
The 'far' subset has more instances of localization failures. As shown in Fig. \ref{fig:result_nearfar}, both the first and second cases of the 'far' subset fail because the depth information of the object is not predicted correctly. Targets with greater depth present a challenge for our task.

\begin{figure*}[h]
  \centering
  \includegraphics[width=0.99\textwidth]{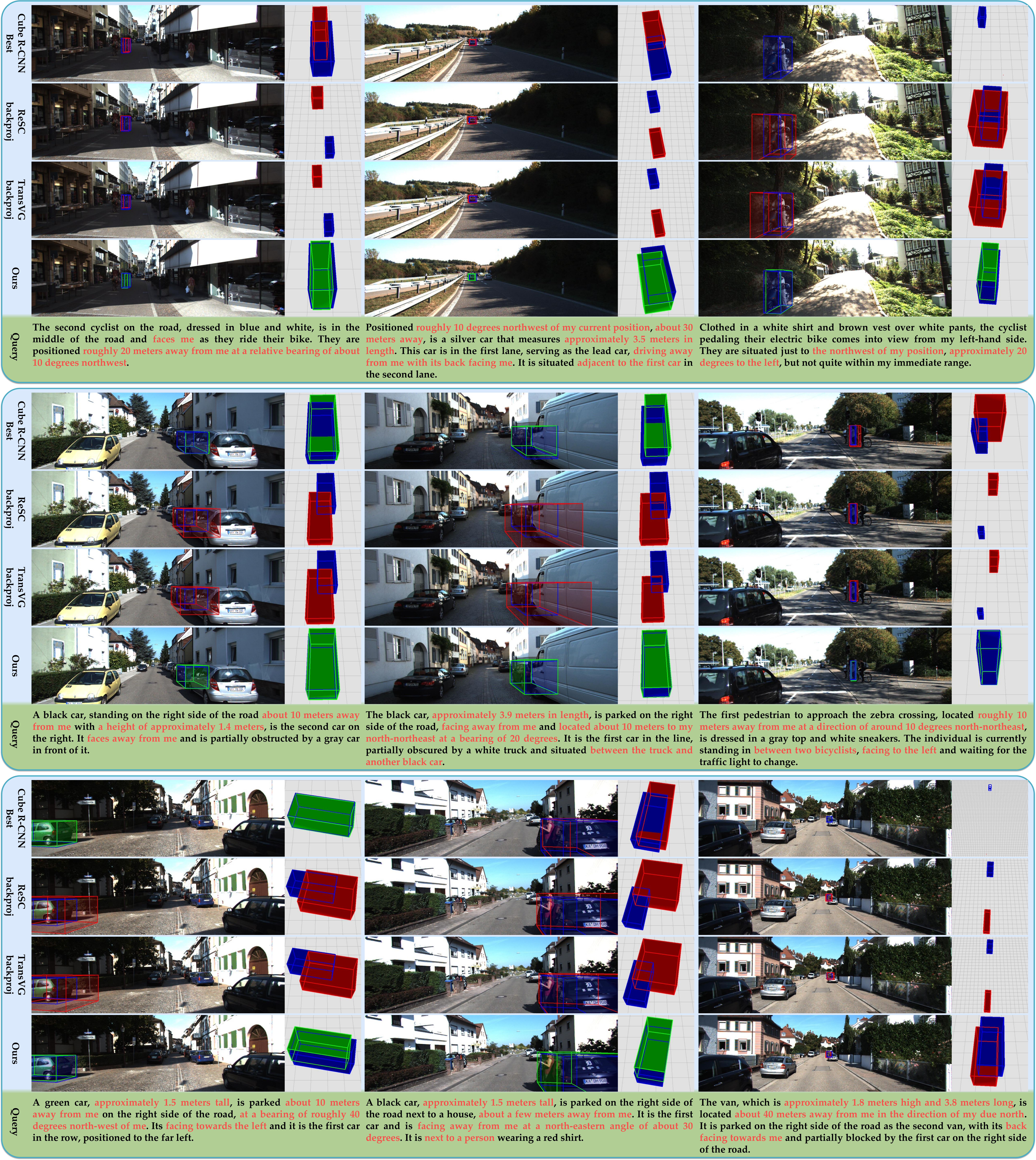}
  \caption{Additional qualitative results in the 'easy', 'moderate', and 'hard' subsets.}
  \label{fig:result_easyhard}
\end{figure*}

\textbf{More qualitative examples in the 'easy', 'moderate', and 'hard' subsets.}
The three blocks in Fig. \ref{fig:result_easyhard} from top to bottom represent the results of 'easy', 'moderate', and 'hard' subsets.
Cube R-CNN has good capabilities in detecting occluded and truncated objects.
Except for the third and ninth cases which are not detected, Cube R-CNN Best either accurately locates the objects or provides approximate 3D ranges, such as the first, second, sixth, and eighth cases.
ReSC backproj and TransVG backproj are able to provide approximate 3D range when the objects are close, such as the third, fourth, fifth, seventh, and eighth cases. However, other objects that are farther away without accurate depth information will produce large deviations.
Fig. \ref{fig:result_easyhard} shows that our method can provide accurate localization for the objects without or with occlusion and truncation.

\begin{figure*}[h]
  \centering
  \includegraphics[width=0.99\textwidth]{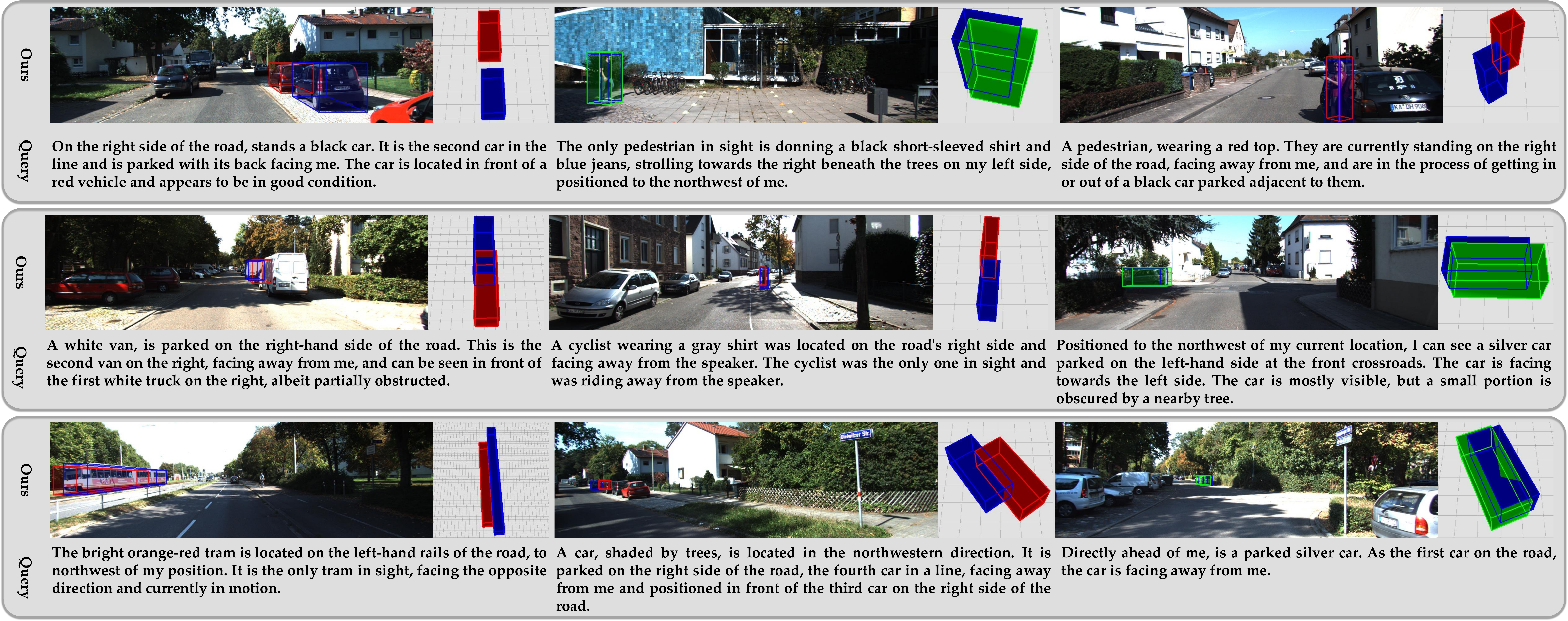}
  \caption{ Qualitative results for the traditional query without specific geometry information corresponding to Fig. \ref{fig:result_nearfar}.
  }
  \label{fig:result_nearfar2}
\end{figure*}

\begin{figure*}[h]
  \centering
  \includegraphics[width=0.99\textwidth]{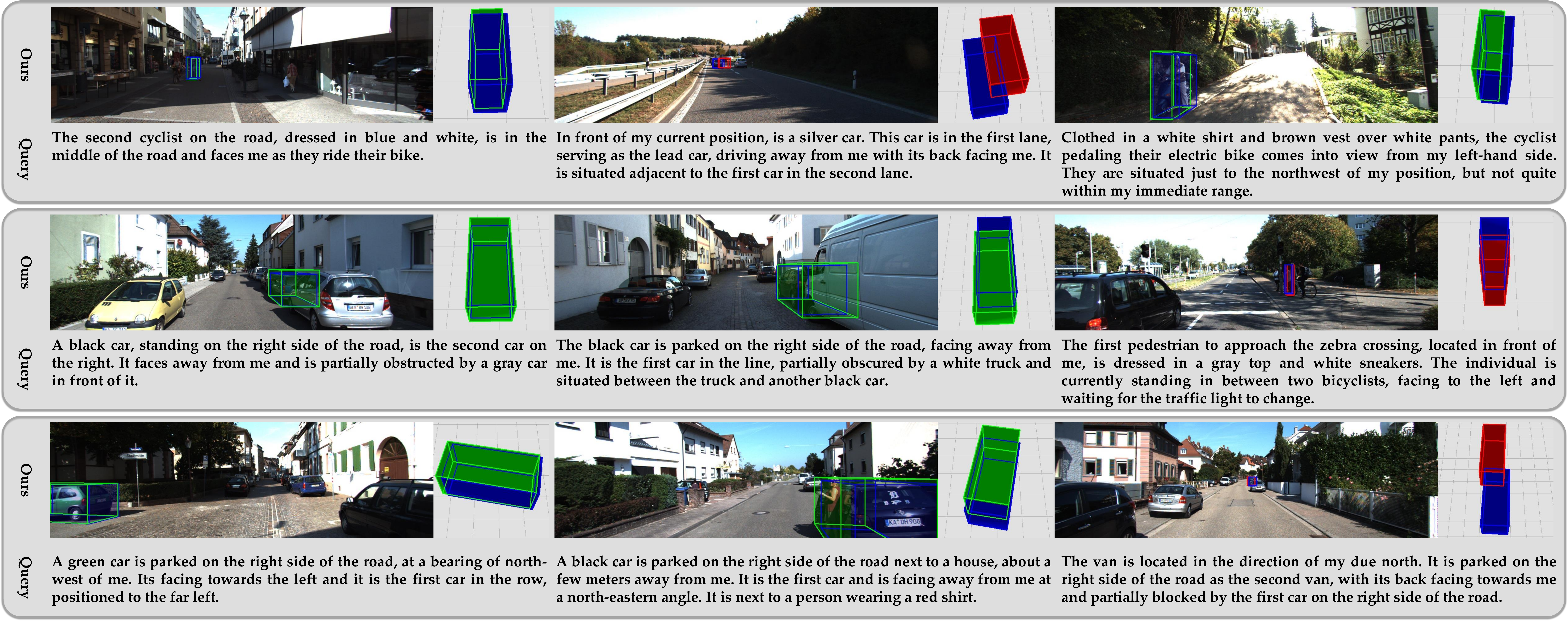}
  \caption{ Qualitative results for the traditional query without specific geometry information corresponding to Fig. \ref{fig:result_easyhard}.
  }
  \label{fig:result_easyhard2}
\end{figure*}

\textbf{More qualitative examples for the traditional query without specific geometry information.}
Fig. \ref{fig:result_unique2} shows the results corresponding to Fig. \ref{fig:result_unique}. From the 'unique' subset, queries without specific geometric information can still provide a rough 3D range. However, for the 'multiple' subset, it is very easy to generate ambiguity and lead to localization failures. 
Fig. \ref{fig:result_nearfar2} presents the results corresponding to Fig. \ref{fig:result_nearfar}. Whether in the 'near', 'medium', or 'far' subsets, the lack of specific geometric information in the text leads to errors in depth estimation and localization failures for the referred objects. 
Fig. \ref{fig:result_easyhard2} displays the results corresponding to Fig. \ref{fig:result_easyhard}. Regardless of the 'easy', 'moderate', or 'hard' subsets, the text query without specific geometric information each has a case of localization failure.

\section{Acknowledgments}
This work was supported in part by grants from the National Science Fund for Distinguished Young Scholars (No.61825603), the National Key Research and Development Project (No.2020YFB2103900), and the Innovation Foundation for Doctor Dissertation of Northwestern Polytechnical University (No.CX2023030).

\bibliography{aaai24}

\end{document}